\documentclass[12pt]{article}
\usepackage{mayahq}
\usepackage[utf8]{inputenc}
\usepackage{graphicx}
\usepackage{pdflscape}
\usepackage{amsmath}
\usepackage{float}
\usepackage{amssymb}
\usepackage{scrtime}
\usepackage[dvipsnames]{xcolor}
\usepackage{hyperref}
\usepackage{fnpct}
\usepackage{titletoc,tocloft}
\usepackage[backend=biber,style=numeric,sorting=ynt,natbib=true]{biblatex}

\hypersetup{
	colorlinks=true,
	urlcolor=MidnightBlue,
	citecolor=PineGreen,
	linkcolor=MidnightBlue,
}

\graphicspath{ {images/} }

\newcommand{\gindex}{\textbf{g-index}~}
\newcommand{\repo}{\href{https://github.com/mayahq/g-index-benchmark}{https://github.com/mayahq/g-index-benchmark}}
\newcommand{\supplus}{$^+$~}
\newcommand{\supminus}{$^-$~}

\title{Towards A Measure of General Machine Intelligence}

\author{
	Gautham Venkatasubramanian\thanks{
		Contributions: Gautham V. and Sibesh K. led the research.  Abhimanyu S. conducted the
		experiments. Shubham M., Dushyant Y. and Shreyansh C. worked on program design and the training dataset.  Thanks
		to Nilay Savant for the DAG visualization diagrams.
	}\\ Maya Labs \\
	\texttt{gautham@mayalabs.io} \\ \and Sibesh Kar$^{\ast}$\\ Maya
	Labs\\ \texttt{sibesh@mayalabs.io} \\ \AND Abhimanyu Singh \\ Maya Labs \\
	\texttt{abhimanyu@mayalabs.io} \\ \and Shubham Mishra\\ Maya Labs\\
	\texttt{shubham@mayalabs.io} \\ \and Dushyant Yadav\\ Maya Labs\\
	\texttt{dushyant@mayalabs.io} \\ \And Shreyansh Chandak\\ Maya Labs\\
	\texttt{shreyansh@mayalabs.io} \\
}

\date{\today}

\begin{document}

\maketitle

\begin{abstract}

	To build general-purpose artificial intelligence systems that can deal with unknown
	variables across unknown domains, we need benchmarks that measure how well these systems
	perform on tasks they have never seen before. A prerequisite for this is a measure of a
	task's generalization difficulty, or how dissimilar it is from the system's prior
	knowledge and experience. If the {\it skill} of an intelligence system
	in a particular domain is defined as it's ability to consistently generate a set of
	instructions (or programs) to solve tasks in that domain, current benchmarks do not
	quantitatively measure the efficiency of acquiring new skills, making it possible to
	brute-force skill acquisition by training with unlimited amounts of data and compute
	power. With this in mind, we first propose a common language of instruction, a
	programming language that allows the expression of programs in the form of directed
	acyclic graphs across a wide variety of real-world domains and computing platforms. Using
	programs generated in this language, we demonstrate a match-based method to both score
	performance and calculate the generalization difficulty of any given set of tasks. We use
	these to define a numeric benchmark called the generalization index, or the \gindex, to
	measure and compare the {\it skill-acquisition efficiency} of any intelligence system on a set
	of real-world tasks. Finally, we evaluate the suitability of some well-known models as
	general intelligence systems by calculating their \gindex   scores.

\end{abstract}

\newpage
\tableofcontents

\section{Introduction}

\subsection{History of defining intelligence}

The concept of intelligence has been expressed in informal terms from time immemorial,
but to date there has been no consensus on a formal definition
\citep{GOTTFREDSON199713}. In psychology, definitions of intelligence include
\textit{``the faculty of adapting one's self to circumstances"} \citep{binet1961development}, and \textit{``the aggregate or global capacity .
	.. to act purposefully, to think rationally, and to deal effectively with [the] environment"},
\citep{wechsler1944measurement}. The definition of artificial intelligence observes similar
variety, with common reliance on a human reference; for eg. \textit{``machines capable of performing tasks that would require intelligence if done by
	humans''}
\citep{quillian1968semantic, minsky1982semantic}. Developments in the field of AI have led to further
refinements, with terms such as skill-based, narrow, or weak AI versus general or strong
AI \citep{Pennachin2007, searle1980minds, Searle:2009}.

The Turing test ``imitation game'' \citep{turing1950computing}, one of the first measures
of artificial intelligence was qualitative in nature. It required that an artificial
intelligence convince a human judge that it was a human. This created an inaccurate
perception of an AI's capabilities due to variance in the judge's knowledge
\citep{dennett1998brainchildren}. The improvements to the Turing test (such as the Lovelace
test \citep{bringsjord2003creativity} and its successor \citep{riedl2014lovelace}) maintain
the requirement of a human or human-like judge, focusing on the judge's available
resources and formal descriptions of what the judge can use for measuring the AI's
capabilities. Till date, subjective human judgment plays a key role in evaluating any
intelligence system.

\subsection{Intelligence as benchmarks of skill}
\label{sub:intelligence-bench}

Recent breakthroughs in the capabilities of machine intelligence systems have relied upon
the scaling of computational methods such as support vector machines, random forests and
neural networks. While this was assisted by the increased availability of raw computing
power, we note that framing machine intelligence in terms of computation enabled
quantitative methods for evaluation, as measuring the performance of a system was
simplified to computing a numerical benchmark score on a publicly available dataset. The
MNIST dataset \citep{mnist} served as a benchmark for digit classification
\citep{oliveira2004support,
	manjunath2007unconstrained, keysers2007comparison} , and is used today in introductory texts to showcase the
power of deep neural networks. Over the years, designing a good benchmark has developed
into a specialized problem, involving the collection of large diverse datasets spanning
multiple years \citep{beyer2020}.

\begin{figure}[htpb]
	\centering
	\includegraphics[width=0.8\linewidth]{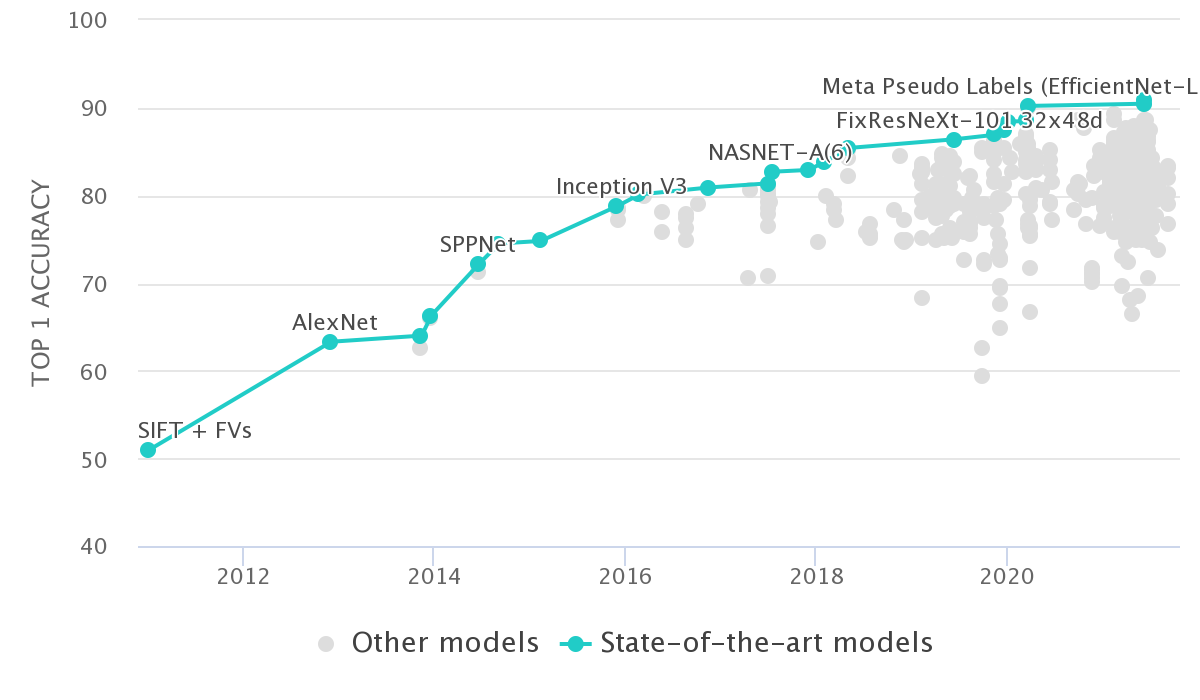}
	\caption{The state of the art in image classification--in context of the Top-1 accuracy benchmark
		on the ImageNet dataset--has improved almost every year since 2012. The presence of a
		consistent numeric benchmark spurred the development of better systems for image
		classification. Image source : \href{https://paperswithcode.com/sota/image-classification-on-imagenet}{https://paperswithcode.com/sota/image-classification-on-imagenet}.
	}
	\label{fig:imagenet}
\end{figure}

The presence of quantitative benchmarks was a boon for developing machine learning-based
intelligence systems, as one could compare different methods and design targeted
improvements to build upon a consistent, computable, numerical score. The
state-of-the-art in image classification improved every year since the ImageNet benchmark
\citep{deng2009imagenet} became widely available, starting with AlexNet in 2012
\citep{krizhevsky2012imagenet}, to SENet \citep{hu2019senet} in 2017 (see
\autoref{fig:imagenet}), with new techniques such as ResNet skip-connections gaining
prominence on the back of their ImageNet performance. Similar improvements were also seen
in the field of language understanding, with the GLUE benchmark
\citep{wang2018glue} and the development of well-known transformer models like
BERT \citep{devlin2018bert} and GPT-2 \citep{radford2019language}.

\subsection{Measuring general intelligence}
\label{sub:measuring-agi}

Deep learning-based methods have achieved a high level of skill in specialized tasks, but
it is still hard to quantitatively measure the “generalization capability” of an
intelligence system - or its “ability to handle situations (or tasks) that differ from
previously encountered situations” \citep{chollet2019}. This is due to the
difficulty of measuring the variables involved in current definitions of artificial
intelligence. \citet{legg2007universal} informally define intelligence as measuring an
\textit{``agent's ability to achieve goals in a wide range of environments''}. It further mentions the properties desirable in a measure of
intelligence, such as a formal mathematical definition, applicability across different
methods without bias, and an informative, numerical score to enable comparison across
agents. However, an agent's intelligence cannot be computed practically according to this
definition as it relies on finding the Kolmogorov complexity \citep{li2008introduction}
of each environment .

\citet{orallo2017} bifurcates the measurement of
intelligence systems into task-oriented and ability-oriented evaluations. Both are
important for evaluating a system, but the former is far more common. Task-oriented
evaluations include human judgments of AI performance, direct benchmark comparisons, and
assessment of adversarial situations in games such as playing Chess or Go.
Ability-oriented evaluations take the form of psychometrics in the case of human
intelligence, and extend to artificial systems \citep{dowe2012iq} when
intelligence is viewed as a form of information processing \citep{chandrasekaran1990}.
This allows the use of algorithmic information theory (AIT) \citep{chaitin1982godel}
to perform ability-oriented evaluations of artificial systems. Building on this,
\citet{chollet2019} provides an outline for the measurement of intelligence via a
framework that is easily mapped to current methods in machine learning. The intelligence
of a system here is defined as \textit{``the measure of its
	skill-acquisition efficiency over a scope of tasks, with respect to priors, experience,
	and generalization difficulty''}, and involves testing via a
benchmark dataset called the Abstract Reasoning Corpus (ARC). However, this framework
does not offer a quantitative measure of generalization difficulty, and all evaluation is
close-ended and binary.

In this paper, we define the \textbf{generalization index}, abbreviated to
\textbf{g-index}, a quantitative benchmark to measure the intelligence of an
artificial system as a computable, numerical value. The naming is inspired from the
\textbf{g-factor}, which is a measure of general ability in the field of
psychometrics \citep{jensen1999g}. It accounts for performance, generalization
difficulty and sample efficiency across a wide range of real-world tasks.
\autoref{sec:factors} describes the experimental setup for the benchmark, and
showcases the components that enable calculating numerical values for evaluation.
\autoref{sec:definition} formally defines the parameters on which the \gindex depends,
constructs a mathematical formula, and shows how the properties of the \gindex follow the
guidelines in current literature. \autoref{sec:expt} evaluates some well-known
transformer models as candidates for a general intelligence system by computing their
\gindex scores, and provides a sample dataset of real-world tasks and their associated
skill programs that can be used with the \gindex \footnote{\repo}.
\autoref{sec:flatland} adapts the \gindex for a toy environment with a simpler
program space. Finally, \autoref{sec:conclusion} notes possible directions for
improvement in the current design and of the \gindex.

\section{Setting up the evaluation}
\label{sec:factors}
\subsection{Components of the \gindex}
\label{sub:evalsetup}

In this section, we describe the details of the components required to compute the
\gindex benchmark. We follow the terminology from \citet{chollet2019}, as it is
easily mapped to supervised learning and reinforcement learning. The description of the
evaluation setup can be given as follows:

\begin{itemize}
	\item A \textit{task} $T$ is specified to an
	      \textit{intelligence system} $IS$.
	\item The intelligence system generates a \textit{skill program} $P$
	      to solve the task. The intelligence system has been trained on a training set (or a
	      \textit{curriculum} $C$) of tasks that may or may not be
	      related to $T$.
	\item A \textit{scoring function} evaluates the \textit{responses} of the skill program
	      $P$ against possible \textit{situations} of the task, and
	      provides a \textit{score} along with some \textit{feedback} if
	      available.
	\item The intelligence system can be updated based on the evaluation and feedback of the
	      scoring function.
	\item The \textit{generalization difficulty} $GD(T, C)$ of a task
	      $T$ measures how different $T$ is from the
	      curriculum $C$ of the intelligence system. It can be used to
	      weight the system's performance for varying degrees of unseen task specifications.
\end{itemize}

While previous definitions of general intelligence rely on quantities like Kolmogorov
complexity \citep{li2008introduction} which are difficult to compute, the components in
our setup together enable computing numerical values that can be combined to measure the
capabilities of an intelligence system. The skill program $P$ is
expressed in a custom programming language that can be extended to construct new programs
without additional syntax complexity, which streamlines collecting and augmenting data
for the intelligence system. The scoring function does not require running the program to
compute the score, which means it can also be used as a loss function or reward function
for training the intelligence system. Finally, we construct an intuitive formulation of
generalization difficulty based on nearest neighbors that reuses the scoring function.
These features are described in the subsections that follow.

\subsection{The Task Specification}
\label{sub:task}

Human beings follow a sequence of steps to perform a specified task. For an artificial
intelligence system, the equivalent sequence of steps is the program. Hence to perform a
task, the system would need to generate (or \textit{synthesize}) programs, when
provided a specification via examples of inputs, in natural language, images, audio, or
video.

The field of \textit{program synthesis} deals with the automatic construction of
programs that are consistent with a given task specification \citep{waldinger1969}.
The common form of a task specification is a set of input-output examples
\citep{amarel1970, summers1977} from which the necessary program(s) can be synthesized. The
application of deep learning to program synthesis is called \textit{neural program synthesis} .
Many neural program synthesis techniques have the task specified via a set of
input-output examples \citep{parisotto2016neurosymbolic} , but some also use natural language
text prompts \citep{lin2017program}, demonstration videos
\citep{sun2018neural}, or combinations of these as well\citep{tfcoder, shu2021agent}.

In our current setup, the tasks submitted to the intelligence system are specified in
English without listing any input-output examples. When scoring the generated program, an
associated reference program is provided.

\subsection{The Skill Program}
\label{sub:program-space}

The choice of target programming language for synthesis varies across implementations.
\citet{yin2017syntactic} describe the generation of Python code snippets from a given
description. \citet{lin2017program} use recurrent neural networks (RNNs) to produce
shell scripts. Codex \citep{copilot}, which uses a large language model
similar to GPT-3 \citep{brown2020gpt3}, generates entire functions in Python from
documentation strings, with similar capabilities being extended to other common
programming languages. It is also common to use a domain-specific language (DSL). DSLs
for program synthesis may be designed from scratch for a specific purpose
\citep{raza2015compositional}, a restricted subset of a language
\citep{tfcoder,
	gramRLprog} or an extended version of an existing language
\citep{bfplus}.

\subsubsection{Programs as directed acyclic graphs}
In our current setup, the intelligence system synthesizes programs that follow the
flow-based programming (FBP) paradigm \citep{morrison1971, morrison2010}. Flow-based programs
are networks of nodes, each encapsulating a "black box" process, transferring data across
predefined connections. Flow-based programs are expressed as directed acyclic graphs
(DAGs) consisting of nodes with various attributes, that are designed to be reusable and
wired together to perform any task.  The exact syntax of flow-based programs can vary,
but can usually be converted into an order-independent array in Javascript Object
Notation (JSON \citep{crockford2012json, pezoa2016foundations}), a data interchange format derived from
Javascript. Node-RED \citep{nodered} and NoFloJS \citep{noflo}
are popular Javascript packages that enable writing flow-based programs, and both allow
for the programs to be saved as JSON. There are several advantages to have the
intelligence system synthesizing programs in the flow-based paradigm:

\begin{itemize}
	\item \textbf{Human-friendly and machine-friendly skill programs}: Flow-based program JSONs can be displayed in a visual
	      programming interface where one can see all the nodes, their connections, global program
	      configuration and possible errors. This makes it easy for humans to construct, edit, and
	      interpret flow programs. The key-value syntax of JSON imposes constraints that make it
	      less complicated than full languages like Javascript and Python, which helps when
	      synthesizing the program DAG. \autoref{fig:ndr1} provides an example of a flow
	      program along with the JSON specification of selected nodes.
	      \begin{figure}[hbt]
		      \hspace*{-0.7in}
		      \centering
		      \includegraphics[width=1.2\linewidth]{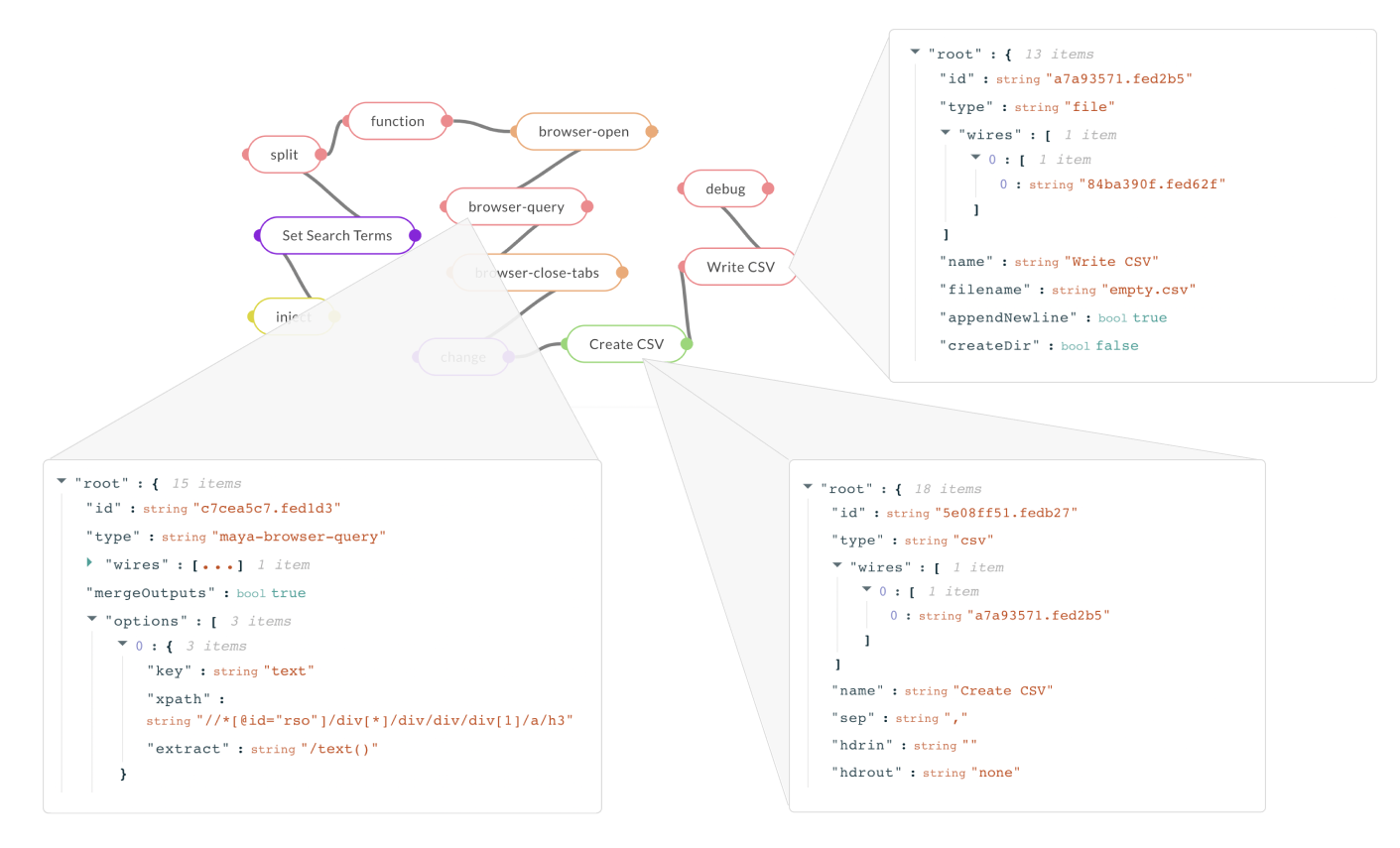}
		      \caption{
			      Representing a flow-based program as a DAG: The JSON descriptions of three nodes are
			      displayed - each node instance has unique \texttt{id} for reference and
			      \texttt{wires} connecting it to other nodes. For the
			      \texttt{maya-browser-query} type node, the attributes include \texttt{mergeOutputs} and \texttt{options}. The
			      \texttt{csv} type node has a wire connecting it to the
			      \texttt{file} node via its unique \texttt{id}.
		      }
		      \label{fig:ndr1}
	      \end{figure}

	\item \textbf{Extensible via custom nodes}: Every node in a flow-based program is
	      assigned a special \texttt{type} attribute, and all nodes with the same
	      \texttt{type} contain the same attribute keys. Most FBP implementations
	      provide a default library of node types for constructing programs, but we can also create
	      and add new node types that encapsulate a custom functionality for our own
	      use\footnotemark. This allows for maximum extensibility and reusability within the same
	      level of expressivity: adding a new node type increases the number of possible skill
	      programs without changing the language syntax or bloating program size.

	      \footnotetext{For examples of reusable nodes see the library in \autoref{sec:node-types}.}

	\item \textbf{Perform variety of tasks}: Each node JSON description in the FBP is an abstraction
	      linked to a particular black box process. The functionality encapsulated in each node can
	      be implemented in any programming language across platforms. This means that flow-based
	      program can be deployed on desktop computers, on servers in the cloud, and even on
	      embedded devices such as the Raspberry Pi and Arduino. This allows us to specify tasks
	      that may be performed across multiple devices, locally or over the internet.
	      \autoref{fig:ndr2} shows four different tasks with their associated flow-based
	      skill programs.
	      \begin{figure}[hbtp]
		      \centering
		      \vspace*{-0.3in}
		      \includegraphics[width=0.8\linewidth,keepaspectratio]{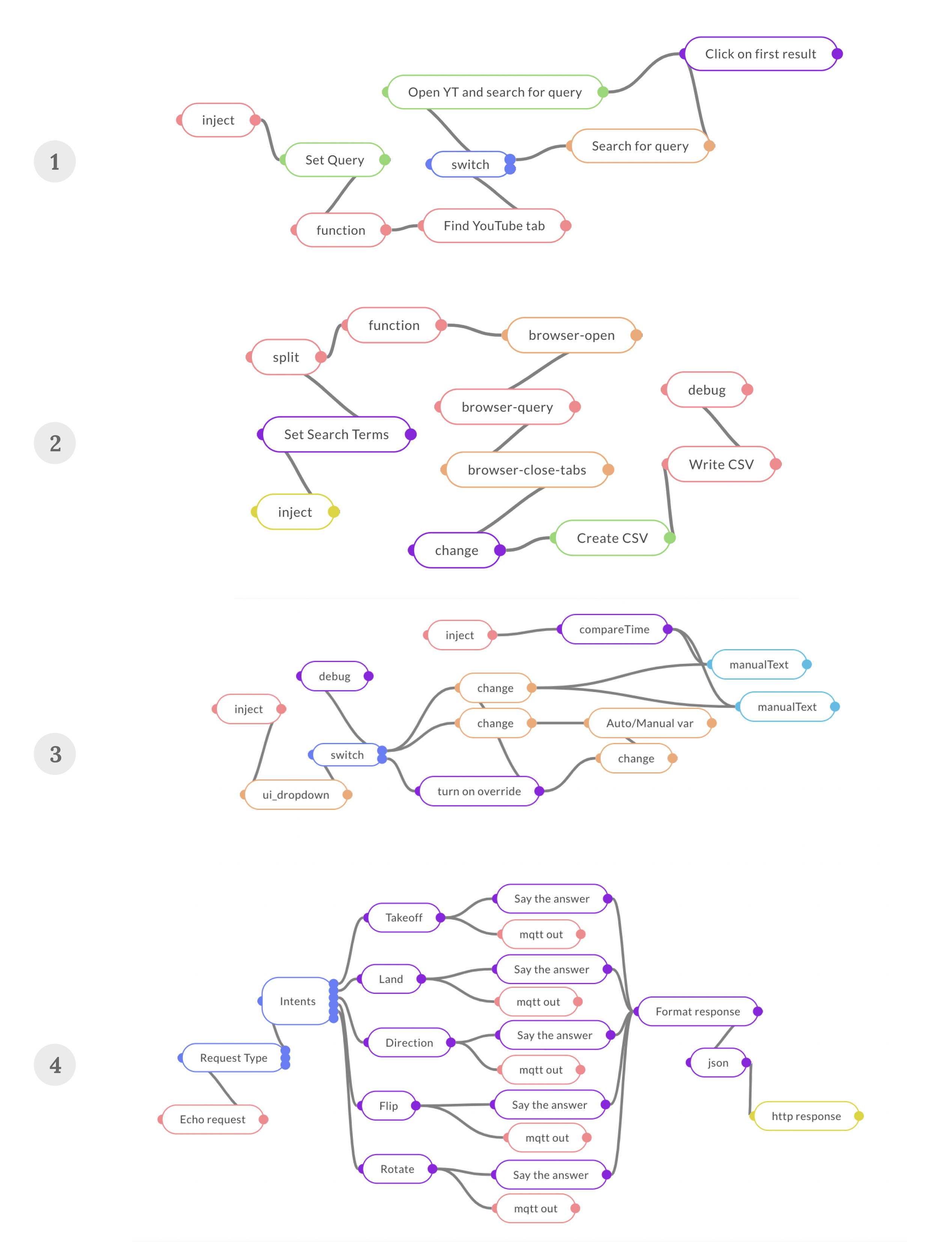}
		      \caption{Program DAGs to : 1. Open browser tab, search for and play a video on Youtube. 2. Search for a given query in a browser using
			      Google Search, scrape the results and save them to a file. 3. Show a dashboard for controlling an
			      Industrial IOT setup and 4. Remotely control a drone using MQTT messages}
		      \label{fig:ndr2}
	      \end{figure}

	\item \textbf{Constrained program design}: The design of flow-based programs can be restricted in three
	      ways:

	      \begin{enumerate}
		      \item The limited syntax of JSON and the DAG construction makes it difficult to use advanced
		            programming constructs like recursion and loops.
		      \item If necessary, the library of available nodes can be customized to prevent the
		            intelligence system from using specific nodes.
		      \item If certain node properties are confusing or risky (such as allowing arbitrary code
		            execution), they can be restricted by designing new nodes that do not allow such
		            modifications.
	      \end{enumerate}

	      Proper restrictions along these axes can limit \textit{program aliasing} -- the
	      existence of multiple valid programs that satisfy the task specification -- by providing
	      one obvious direction to solve a given task.

	\item \textbf{Efficient program generation} - Program synthesis methods based on deep
	      learning may have limits on the size of synthesized programs.	For instance, neural
	      network architectures like transformers have a fixed upper bound on the number of tokens
	      that can be generated, so it is important to use the available token space efficiently.
	      With flow-based programs, we can succinctly specify nodes that perform complex tasks due
	      to the power of encapsulation. This allows for a larger space from which the intelligence
	      system can generate programs.

	\item \textbf{Evaluating a generated program}: The programs synthesized by the
	      intelligence system are evaluated by the scoring function.  It is impractical to evaluate
	      the outputs of the skill program against all possible inputs, so approximate techniques
	      are used, such as comparing program structure or using special input cases. Programs
	      written in Javascript or Python may be difficult to evaluate without testing due to
	      program aliasing. Since program aliasing can be minimized for flow-based programs, they
	      can be evaluated without having to run the program.

	\item \textbf{Language-agnostic skill programs}: The nodes used in flow-based programs
	      encapsulate black-box processes, which means that the underlying implementations can be
	      in any programming language. Flow-based programs hence simply act as a coordination layer
	      between implementations of different pieces of logic. This means that models that learn
	      how to synthesize these DAGs only need to learn the abstract relationship between task
	      input and its solution program, leaving the low-level details free to be implemented in
	      any manner. \autoref{fig:agnostic} shows flow-based programs across different tasks.

	      \begin{figure}[hbtp]
		      \centering
		      \vspace*{-0.3in}
		      \hspace*{-0.55in}
		      \includegraphics[width=1.3\linewidth,keepaspectratio]{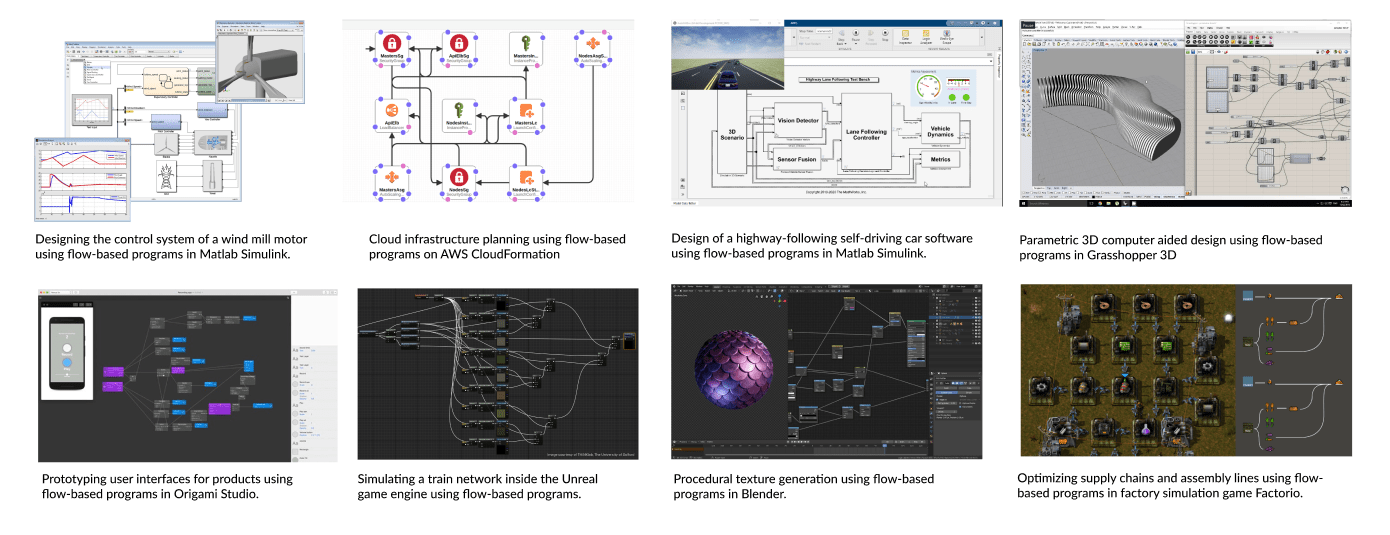}
		      \caption{Since flow-based programs are simply the coordination logic to compose any set of
			      processes together, these process components can be built using any framework, language, application or API, leading to a wide range of applications. }
		      \label{fig:agnostic}
	      \end{figure}
\end{itemize}

\subsubsection{DAGs vs Abstract Syntax Trees}
\label{dags-vs-asts}

While programs in other programming languages can be analysed as graphs by examining
their Abstract Syntax Trees (AST) representations, the DAGs of flow-based programs are
easier to compare with one another due to the following reasons:

\begin{itemize}
	\item The order of nodes in the AST is dependent on the order of text in the program source.
	      For our program DAGs, the order of the program is specified implicitly in the program
	      source by the edges between nodes. DAG comparisons are less affected by the order of
	      information specified in the program source compared to AST comparisons.
	\item AST representations suffer from program aliasing--two programs that satisfy the same task
	      can have completely different ASTs. For our program DAGs, we can minimize program
	      aliasing by constraining the kinds of nodes that are allowed for use.
	\item For complex specifications, the program size (and therefore AST size) can grow
	      arbitrarily large if encapsulation is not used, which makes it harder to compare two
	      given programs. Our program DAGs are designed to benefit from encapsulation: each node
	      can encapsulate a ``black-box'' process of arbitrary complexity, and the attributes of
	      the node are used to examine and control the behavior of the process.
	\item The AST is a low-level representation of the program: it is used to ensure that the
	      program source is syntactically valid, check for minor semantic errors (like
	      dereferencing null pointers) and perform program optimizations. It is difficult to reason
	      about the behavior of the program from looking at its AST representation. The DAG is a
	      high-level representation of the flow-based program: we can infer the program's general
	      behavior from the DAG structure and the node types, and examine node attributes to
	      understand or change the behavior of any component.
\end{itemize}

\subsection{The scoring function} \label{sub:scoring}

After a skill program has been synthesized by the intelligence system, it is evaluated by
a scoring function.  The program can be evaluated by its success on a special set of
input-output pairs, or by match-based metrics. For a given task, match-based metrics
compute similarity by comparing the structure a generated program with a known reference
program. This reference could be provided by a human, generated via a fixed set of rules,
augmented from existing data, or synthesized by another intelligence system. The BLEU
score \citep{bleu2002score} can be used to compare the text of the two given
programs, but it does not consider the structured syntax or the semantic features of the
programs.  CodeBLEU \citep{ren2020codebleu} improves upon the BLEU score by comparing
the abstract syntax trees (AST) and the semantic dataflow of the programs. While
match-based metrics do not need to run the generated program for evaluation, they are
affected by program aliasing. Recent synthesis methods \citep{kulal2019spoc,lachaux2020unsupervised,copilot}
evaluate programs via \textit{functional correctness}, wherein a generated program is
considered correct if it passes a set of unit tests\footnotemark. Functional correctness
is useful because it is similar to how humans evaluate programs written by each other,
but it requires running the generated program to obtain a score.

\footnotetext{Unit tests include a set of known input-output pairs, but may also contain collections of inputs that together test for the presence of certain
	properties such as types of failures in a given program.}

\subsubsection{Divergence Metric for Flow-Based Programs}
\label{ssub:div-define}

In \autoref{sub:program-space} we noted that flow-based programs can be constrained to
minimize the program aliasing, and the DAG representations are easier to compare than
ASTs (\autoref{dags-vs-asts}). Hence for our scoring function, we use a match-based
divergence metric $\Delta$ to compare the DAG of the generated program
with a known reference.

\begin{itemize}
	\item Let $P'$ be a known reference skill program, and
	      $P''$ be a skill program generated by the intelligence system.
	\item Let $G'(V', E')$ be the DAG denoting the $P'$. Here,
	      $V'$ and $E'$ refer to the vertices and edges
	      of $G'$ respectively, and $E' \subset V' \times V' $.
	\item Let $G''(V'', E'')$ be the DAG denoting the program
	      $P''$. Here, $V''$ and
	      $E''$ refer to the vertices and edges of $G''$
	      respectively, and $E'' \subset V'' \times V'' $.
\end{itemize}

Given two programs $P', P''$, the divergence metric
$\Delta$ accepts their DAGs $G', G''$ as input and
is constrained as follows:

\begin{equation}
	\label{metric-constraints}
	\begin{aligned}
		 & \text{$\Delta$~is bounded}   & \implies &  &  & \Delta(G',G'') \in [0, 1]          ~\forall~\text{DAGs}~~G', G'' \\
		 & \text{$\Delta$ is symmetric} & \implies &  &  & \Delta(G', G'') = \Delta(G'', G')  ~\forall~\text{DAGs}~~G', G'' \\
		 & \Delta(G',G'') = 0           & \implies &  &  & G', G'' ~~\text{denote the exact same program $P$}               \\
		 & \Delta(G',G'') = 1           & \implies &  &  & G', G'' ~~\text{do not have anything in common}
	\end{aligned}
\end{equation}

\subsubsection{Computing $\Delta$ for a Single Node}
\label{ssub:div-simple}

Consider the simple case for $\Delta$, where the DAGs
$G'$ and $G''$ both contain only one
attributed node and zero edges. If $v'$ is the only node in
$G'$ and $v''$ the only node in
$G''$, we can compare the attributes of $v'$
with those in $v''$ to obtain a \textit{node similarity} value
$w$.

\begin{itemize}
	\item If both nodes have different \texttt{types}, $w = 0$
	\item If both nodes have the same \texttt{type}, they will have the same attribute
	      keys. Node attribute values follow a binary comparison. If $p$ is
	      number of attributes for a given node \texttt{type}, and
	      $p_{match}$ is the number of attribute values that are equal for both
	      nodes, then $$w(v', v'') = \frac{p_{match}}{p}$$
	\item If both nodes have the same \texttt{type}, and all node properties are
	      equal, $w = 1$.
\end{itemize}

Thus when $G'$, $G''$ both contain only one
node and zero edges, we define $\Delta$ as

\begin{equation}
	\label{metric-simple}
	\Delta(G', G'') = 1 - (w(v', v''))^2
\end{equation}

We can see that the value of $\Delta$ follows the constraints given in
\autoref{metric-constraints} for the simple case.

\subsubsection{Computing $\Delta$ for the General Case}
\label{ssub:div-everything}

When computing the value of $\Delta(G', G'')$ between two arbitrary DAGs, we
would need to consider the \textit{common substructure} between the DAGs in addition to
the node-wise similarity values. The value of $\Delta$ should be low
when the DAGs have highly similar structure, so we compute it by finding the largest
subgraph common to both DAGs. This is known as the \textit{maximum common edge subgraph problem} (MCES)
\citep{bokhari1981mapping}, an extension of the subgraph isomorphism problem
\citep{ullmann1976algorithm}. Two graphs $G'(V', E')$ and
$G''(V'', E'')$ are isomorphic to each oher ($G' \simeq G''$) if
there exists a bijection $\phi: V'  \rightarrow V''$ that preserves the graph structure:

\begin{equation}
	\label{eq:iso-struct}
	(v_1, v_2) \in E' \iff (\phi(v_1), \phi(v_2)) \in E'' ~~~ \forall (v_1, v_2) \in E'
\end{equation}

We need to find subgraphs $ G'_{opt} \subseteq G' $ and $ G''_{opt} \subseteq G'' $
that are isomorphic to each other, ie $G'_{opt} \simeq G''_{opt}$. This requires that
$G'_{opt}$ and $G''_{opt}$ have the same number of
vertices, ie

$$ | V'_{opt} | = | V''_{opt} | $$

and a bijection $\phi_{opt}: V'_{opt} \rightarrow V''_{opt} $ between the subgraphs that satisfies
\autoref{eq:iso-struct}. $G'_{opt}$ and $G''_{opt}$
also need to be the \textit{largest} common subgraphs

\begin{equation}
	\begin{aligned}
		 &  & ~~~~G'_{opt} \simeq G''_{opt}        & ~~                           \\
		 &  & \nexists~G'_{opt2} \simeq G''_{opt2} & \text{~~such that~~}         \\
		 &  & G'_{opt} \subset G'_{opt2},          & G''_{opt} \subset G''_{opt2}
	\end{aligned}
\end{equation}

\begin{figure}
	\centering
	\includegraphics[width=0.9\linewidth]{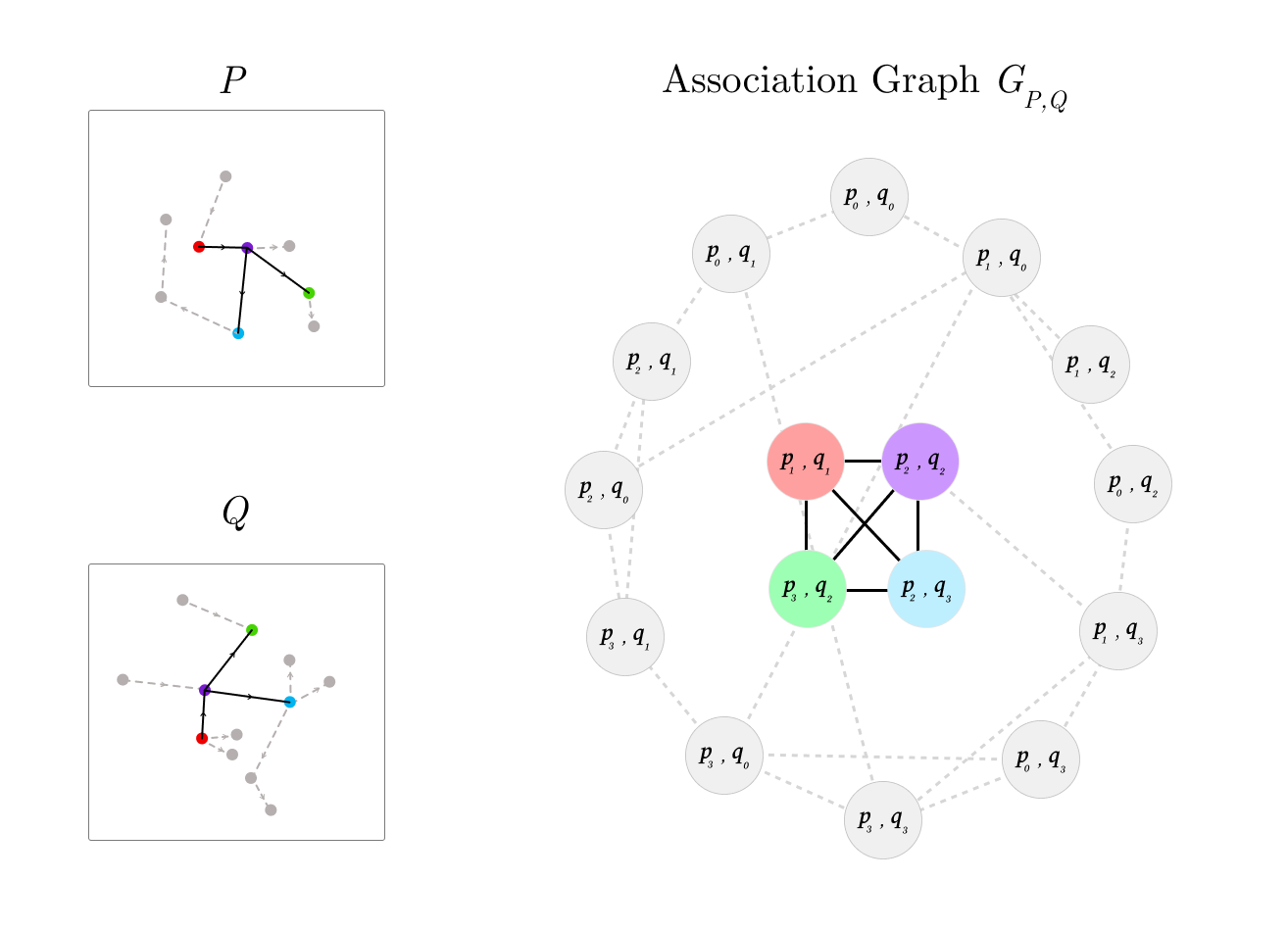}
	\caption{
		\textbf{Computing $\Delta$ for the general case}: For two DAGs with an arbitrary number of nodes and
		edges,
		we compute the value of $\Delta$ by finding the largest common subgraph
		isomorphism $\phi: P \rightarrow Q$ via the association graph
		$G$. In the above figure, a node $(p_i, q_j)$ in $G$ refers to a possible
		association $\phi(p_i) = q_j$. An edge between two nodes in
		$G$ means that the respective associations satisfy
		\autoref{eq:iso-struct}. The largest clique in $G$ (with
		colored nodes and darkened edges) signifies the maximum association between
		$P$ and $Q$ i.e. the largest common
		subgraph. The process is explained in detail in \autoref{sec:delta-actual}.
	}
	\label{fig:divergence-figure}
\end{figure}

We obtain $G'_{opt}$, $G''_{opt}$, and the bijective
mapping $\phi_{opt}$ by constructing an \textit{association graph}
$G$ between the nodes of $G'$ and
$G''$, and finding a maximum clique in $G$
\citep{barrow1976subgraph, kozen1978clique}. \autoref{sec:delta-actual} explains the process of
obtaining $\phi_{opt}$ in detail, including the rules for construction of
the association graph and ensuring the properties of $\Delta$ from
\autoref{metric-constraints} are satisfied. \autoref{fig:divergence-figure} provides a visual
overview of the process.  Once we obtain $\phi_{opt}$, we calculate the
value of the metric $\Delta$ is using each node in
$V'_{opt}$, and its image via $\phi_{opt}$ in
$V''_{opt}$. If $V'_{opt} = \{ v'_1, v'_2, ... v'_N \}$, then

\begin{equation}
	\label{metric-full-glitch}
	\Delta(G', G'') = 1 - \frac{(\sum^{N}_{i=1}w(v'_i, \phi_{opt}(v'_i))^2}{|V'||V''|}
\end{equation}

\subsubsection{Features of $\Delta$}
\label{ssub:div-features}

While $\Delta$ has an exponential complexity due to the use of
subgraph isomorphism, in practice, the computation of $\Delta$ is sped
up as \autoref{assoc-verts} reduces the number of vertices in the association graph
$G$, and \autoref{assoc-edges} tends to produces sparser
graphs. As $\Delta$ does not require the execution of the generated
programs, it can be used in the training loop of supervised or reinforcement learning
methods.

With the match-based DAG divergence metric $\Delta(G', G'')$, we now have a
scoring function to compare a generated flow-based program $P''$
with a reference program $P'$. Flow-based programs can be
expressed in JSON, so a JSON parser can be used to ensure valid syntax. The DAGs
constructed from the valid JSON can then be input to $\Delta$ to
compare the semantic dataflow of the programs. Since $\Delta$ enables
the comparison of any two flow-based programs, it can be extended to measure the
dissimilarity of a program from a given set of programs, and more generally measure the
dissimilarity between two sets of programs, by computing all necessary pairwise
comparisons.

The values of $\Delta$ are based on the structural differences between
the DAGs and consider degrees of errors in the generated program:

\begin{itemize}
	\item \textbf{Syntax Errors}: The generated program $P''$ has
	      incorrect JSON syntax, resulting in a reduced or incomplete DAG after parsing.
	\item \textbf{Function Errors}: Some nodes are incorrectly specified or missing from the
	      generated DAG. If a node in the generated DAG has only one differing attribute compared
	      to the reference, it is reflected in when computing the node similarity
	      $w$.
	\item \textbf{Dataflow Errors}: The generated program has the same nodes as the
	      reference, but has different edges, denoting different semantics.
\end{itemize}

\subsection{Measuring Generalization Difficulty}
\label{sub:measure-gd}

Humans have the capability to \textit{generalize}, or they are able to use past
learning experience to navigate new situations in the present. A machine intelligence
would need to possess similar capabilities to deal with new or unseen task
specifications.  In machine learning, generalization deals with a model's performance on
previously unseen data samples that are similar to the distribution of the training set
of the model. A model is said to have \textit{overfit} the training data if
predicts well on the training data, but fails to predict on new unseen samples. In deep
learning, it is common to use some regularization techniques to avoid overfitting.

In the context of the \gindex and our experimental setup, we focus on the
\textit{generalization difficulty} of a specified task. For a given task
$T'$, \citet{chollet2019} informally defines
generalization difficulty as \textit{``a measure of
	how much the shortest training-time solution program needs to be edited in order to
	become an appropriate evaluation-time solution program''}, and relies on Relative
Algorithmic Complexity to compute the edit difference between the programs. We rephrase
this statement for our experiment setup as follows: Suppose an intelligence system
$IS$, trained on a curriculum of tasks
$C$, is given a task $T'$. If
$P'_{opt}$ is an appropriate evaluation-time program that solves the
task $T'$, how much does $P'_{opt}$ differ from
the training-time solution programs $P_T$ for tasks
$T \in C$?

Since the skill programs generated by the intelligence system in our experimental setup
are flow-based programs, they can be expressed as DAGs. Therefore, if
$P'_{opt}$ is an optimal flow-based program that solves the task
$T'$, we can use the $\Delta$ metric defined in
\autoref{sub:scoring} and quantify the difference between
$P'_{opt}$ and programs generated for tasks in the curriculum
$C$ by a nearest neighbor search. Thus, we can define the
\textit{domain distance} $\Omega$ of a single task
$T'$ for an intelligence system trained on a curriculum of tasks
$C$ as :

\begin{equation}
	\label{domain-distcalc}
	\begin{aligned}
		\Omega(T', C) & = \min\limits_{T \in C} \Delta(G_{P'_{opt}}, G_{P_T})                                          &   & \\
		              & \text{where~}  G_{P'_{opt}} \text{~denotes the DAG of the optimal solution $P'_{opt}$ of $T'$} &   & \\
		              & \text{and~}    G_{P'_T} \text{~denotes the DAG for a program $P_T$ that solves
		$T \in C$}    &                                                                                                &
	\end{aligned}
\end{equation}

Note that $\Omega$ is bounded between $[0, 1]$. If
$\Omega(T', C) = 0$ it means that the task $T'$ can be
found in the training set $C$, and hence no generalization is
required.

\begin{figure}[H]
	\centering
	\includegraphics[width=0.6\linewidth, keepaspectratio]{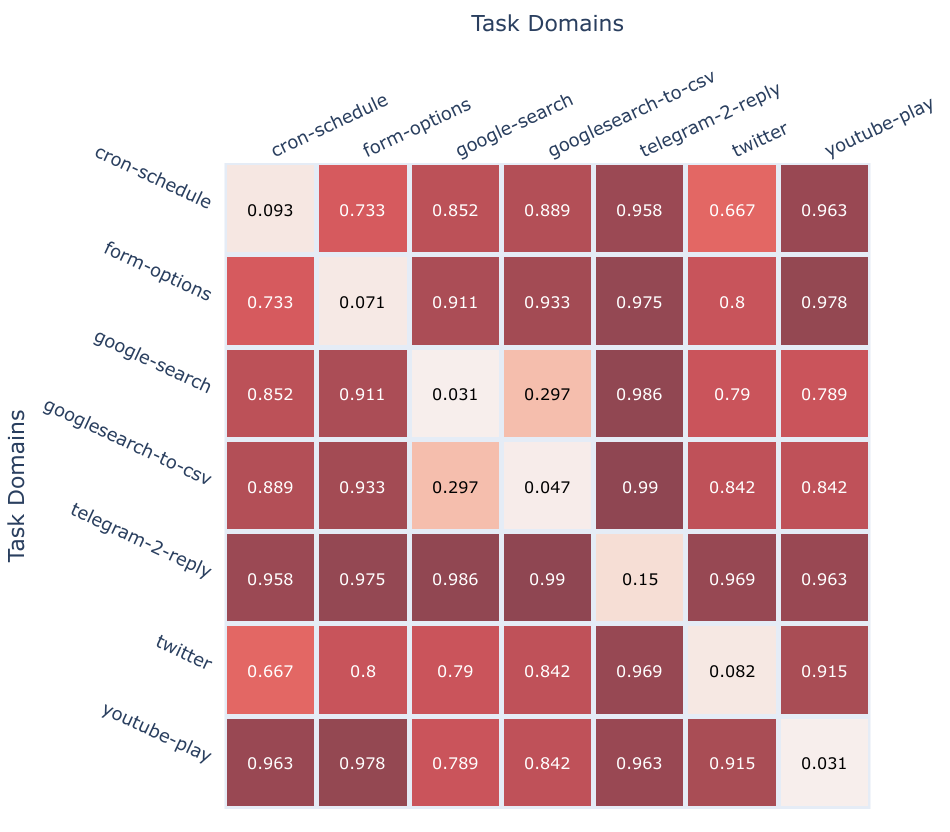}
	\caption{
		Scores of the divergence metric $\Delta$ between skill programs across
		seven different \textit{task domains}: \texttt{cron-schedule, form-options, google-search, googlesearch-to-csv, telegram-2-reply, twitter,} and
		\texttt{youtube-play}. Skill programs within the same domain have scores closer to
		0, and those from different domains have scores closer to 1. A description of task domains
		is given in \autoref{sec:domains-full}.
	}
	\label{fig:domain-distance}
\end{figure}

If $C$ is too large, the value $\Omega(T', C)$ can be
approximated by clustering the elements of $C$ into
\textit{task domains} using the divergence metric $\Delta$: two
tasks within the same domain $C_i$ are ``closer'' to each other
($\Bar{\Delta}\lesssim0.15$) than tasks in different domains.
\autoref{fig:domain-distance} shows an example of task domains and their related divergence
scores.

The definition of $\Omega(T', C)$ in \autoref{domain-distcalc} enables the
computation of generalization difficulty within of the training-set/test-set paradigm
that is commonly used for training machine learning models. By computing
$\Omega$, a model's performance on an unseen dataset can be weighted
in context of the generalization difficulty of the tasks in the set.

\section{The g-index benchmark}
\label{sec:definition}
\subsection{Defining the formula}

We now define the formula for computing the \gindex based on the experimental setup
defined in \autoref{sec:factors}. First, we describe the necessary variables:

\begin{itemize}
	\item An \textbf{intelligence system} $IS$ generates a flow-based
	      \textbf{skill program} $P'$ when input the specification for a
	      given \textbf{task} $T'$. This formulation is
	      method-agnostic, and allows the \gindex benchmark to apply not only for deep-learning
	      based systems of today, but also be extended to any new techniques in the future, by
	      plugging in the variables measured below.

	\item The intelligence system is trained on a \textbf{curriculum}
	      $C$, consisting of tasks $T$ and their
	      associated reference skill programs $P_{T}$. We expect that the ideal
	      intelligence system would also be the most sample-efficient in terms of training -- it
	      would learn to solve a large variety of tasks from just a single demonstration. Hence the
	      value of \gindex should decrease as the number of training samples increases.

	\item A \textbf{curriculum domain} $C_i$, refers to a subset of
	      $C$ where all tasks $T\in C_i$ belong to the same
	      task domain. We expect that an ideal intelligence system would be able to generalize from
	      the least number of tasks provided per domain. So, the value of the \gindex for the
	      system should decrease as more tasks are provided for a given domain. We define
	      $W_{C_i} \in [0,1]$, the weight of considering curriculum tasks from
	      $C_i$ as:

	      $$ W_{C_i} = \frac{1}{1 + log_2(|C_i|)} $$

	\item The \textbf{priors} $\rho$ of the system refer to
	      knowledge built into the system before any training. Examples of priors include
	      previously learnt weights, neural network architectures, data pre-processing techniques,
	      program optimizations etc. We expect that the ideal intelligence system would be able to
	      generalize from having the least amount of built-in priors. Hence the value of the
	      \gindex should decrease as more priors are encoded into the system. For our experiments,
	      we consider the value of $\rho$ as a fixed constant, but we expect
	      this to change as more complex systems are evaluated.

	\item When training the intelligence system $IS$ on a curriculum
	      $C$, it is important to measure the \textbf{experience} of
	      the system with $C$. The units of measure for compute power are
	      FLOPS(Floating Point Operations Per Second) or MIPS(Million Instructions Per Second),
	      which are reflective of the rate at which the system is exposed to the data. We expect
	      that the ideal intelligence system would be one that exhibits high performance after
	      being trained for the least amount of time, with the least amount of compute power.  So,
	      the value of the \gindex should decrease as the intelligence system is trained for longer
	      and on larger amounts of compute power. We define the \textbf{experience}
	      $E$ of the system for a given curriculum
	      $C$ in terms of the compute power used for training
	      $IS$ on $C$ (measured in teraFLOPS)
	      multiplied by the amount of time $IS$ was trained on
	      $C$ (measured in seconds).

	      $$ E(C) = log_2(\text{compute power used on~}C~\cdot~\text{time trained on~}C) $$

	\item A \textbf{scoring function} evaluates the skill program $P'$ for
	      the task $T'$ to measure the performance of the intelligence
	      system $IS$ . We expect that an ideal intelligence system would
	      produce a perfect skill program $P'$ for
	      $T'$. So, the value \gindex for the system should increase as its
	      performance on the scoring function improves. We use the divergence metric
	      $\Delta$ from \autoref{sub:scoring} and compute the difference
	      of the generated skill program $P'$ with a known reference program
	      $P'_{opt}$ to calculate the \textbf{performance}
	      $\theta$ of $IS$ on the task
	      $T'$ as:

	      $$ \theta(IS, T') = 1 - \Delta(P', P'_{opt}) $$

	\item When testing the capabilities of $IS$ after training on a
	      curriculum $C$, we wish to see how $IS$
	      performs on tasks of varying dissimilarity from $C$. We expect
	      that the ideal intelligence system would be able to generalize to tasks that are highly
	      dissimilar from those on which it was trained. Hence the value of the \gindex should
	      increase non-linearly if the system performs well on tasks of increasing
	      \textbf{generalization difficulty}. We define the \textbf{generalization difficulty}
	      $GD$ of a task $T'$ for a system trained on
	      a curriculum $C$ using $\Omega$ defined in
	      \autoref{domain-distcalc} and the exponential function $exp$:

	      \begin{equation}
		      \label{task-gd}
		      GD(T', C) = exp(10 \cdot \Omega(T', C))
	      \end{equation}

\end{itemize}

We use a set of tasks $\{T'_1, T'_2, ...~T'_j,~...\}$ to evaluate an intelligence system
$IS$ trained on a curriculum $C$. The
\textbf{task contribution} $TC$ of a single task
$T'_j$ to the \gindex using the performance
$\theta$, the generalization difficulty $GD$,
the priors $\rho$, and the experience $E$:

\begin{equation}
	\label{eq:task-contr}
	TC(IS, T'_j) = \sqrt{ exp(12 * \theta(IS, T'_j)) \cdot \left [ \sum\limits_{C_i \subset C} W_{C_i} \cdot
			\left ( \frac{GD(T'_j, C_i)}{\rho + E(C_i)} \right ) \right ] }
\end{equation}

The constants and component functions used to calculate each variable's impact are
determined by trends seen during experimentation. Thus we obtain the formula for the
\gindex by averaging over the evaluations for the set of tasks
$\{T'_j\}$:

\begin{equation}
	\label{the-g-index}
	\begin{aligned}
		\text{\gindex}(IS, \{T'_j\}) & = & \frac{1}{|\{T'_j\}|} & \cdot \sum_{T'_j} TC(IS, T'_j)                                                                              &  & \\
		                             & = & \frac{1}{|\{T'_j\}|} & \cdot \sum_{T'_j} \sqrt{ exp(12 * \theta(IS, T'_j)) \cdot \left [ \sum\limits_{C_i \subset C} W_{C_i} \cdot
				\left ( \frac{GD(T'_j, C_i)}{\rho + E(C_i)} \right)
		\right ]}                    &   &
	\end{aligned}
\end{equation}

\subsection{Properties of the \gindex}
\label{sub:g-index-props}

\begin{figure}[htbp]
	\centering
	\includegraphics[width=0.8\linewidth, keepaspectratio]{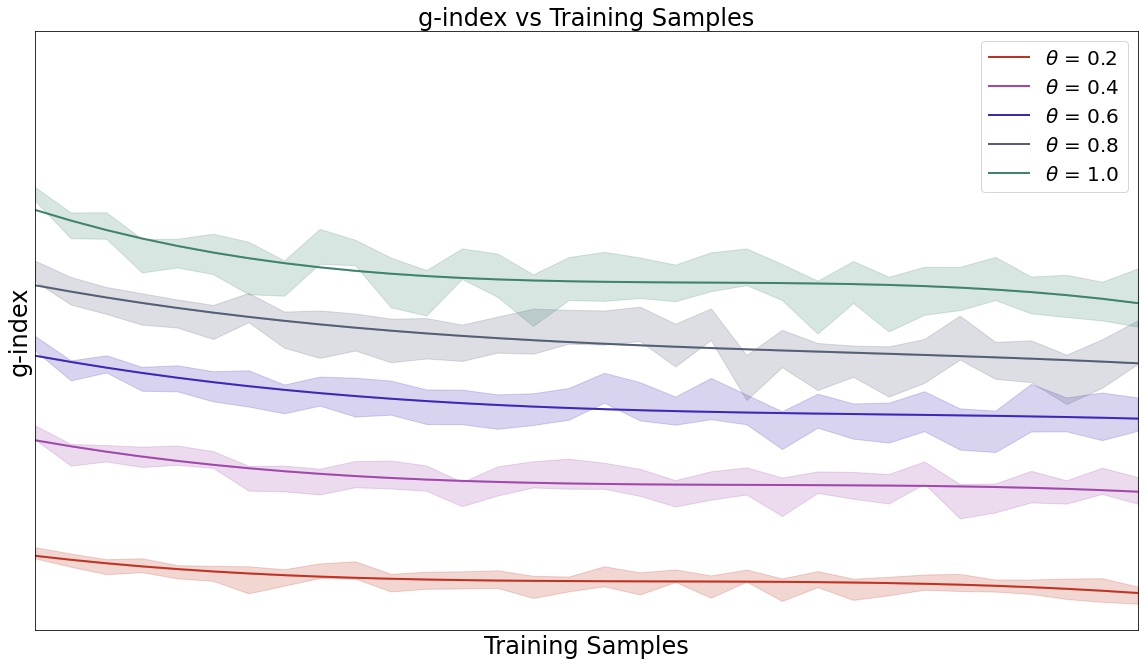}
	\caption{Responsiveness of the \gindex to an increase in the number of training samples (the size
		of curriculum $C$). We assume the priors and experience of the
		system to be constant across all the lines. Each line assumes the performance of the
		intelligence system $IS$ to be fixed at a certain value (shown in
		the legend) and computes the \gindex for the training samples split evenly across all
		task domains. The lightly-shaded region around each line denote the variance of the
		\gindex with respect to the possible uneven distributions across the task domains --
		uneven distributions may have noticeable effect on the \gindex
		value, depending on how difficult the tasks were to generalize. We notice that the \gindex
		steadily decreases as the number of training samples increase. We expect that the ideal
		intelligence system would be closer to the top left of the plot to achieve a high
		\gindex.}
	\label{fig:rando-graphs1}
\end{figure}

\begin{figure}[htbp]
	\centering
	\includegraphics[width=0.8\linewidth, keepaspectratio]{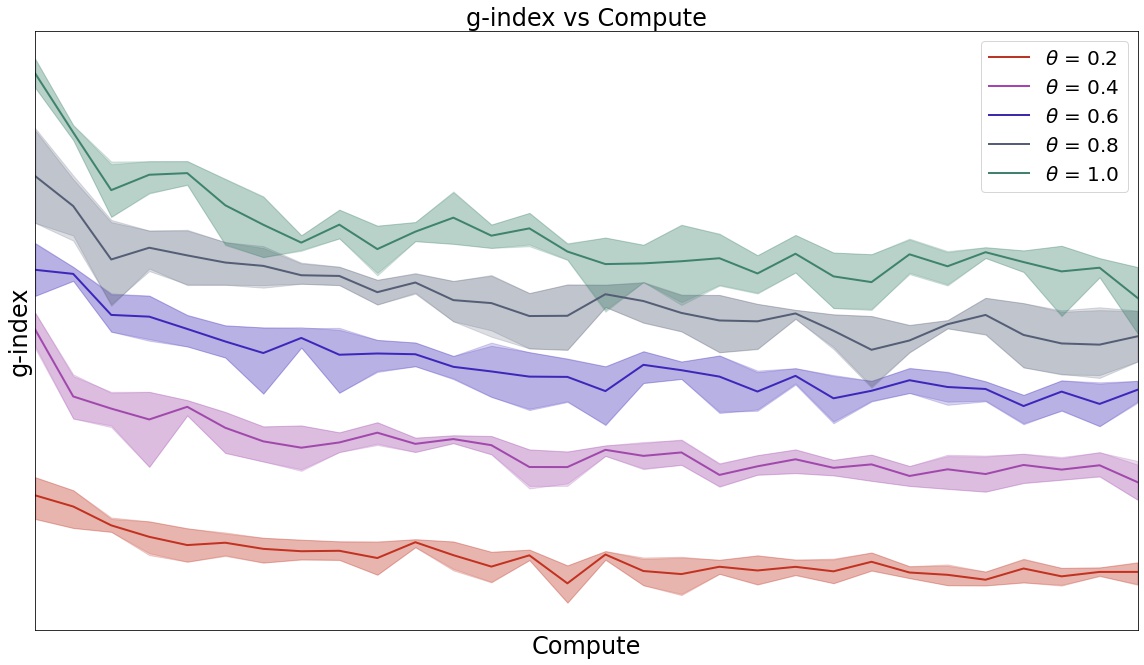}
	\caption{Responsiveness of the \gindex to an increase in available computing power (used in
		computing $E$). We assume the priors of the system to be fixed
		across all the lines. Each line assumes the performance of the intelligence system
		$IS$ to be fixed at a certain value (shown in the legend) and
		computes the \gindex for a given amount compute power, with training samples split evenly across all task domains. The lightly-shaded region
		around each line denotes the variance of the \gindex with respect to possible uneven
		distributions across the task domains -- compared to \autoref{fig:rando-graphs1} uneven
		distributions do not have as high an effect on the \gindex as the compute power
		increases. We expect that the ideal intelligence system would be closer to the top left
		of the plot to achieve a high \gindex.}
	\label{fig:rando-graphs2}
\end{figure}

\begin{figure}[htbp]
	\centering
	\includegraphics[width=0.8\linewidth, keepaspectratio]{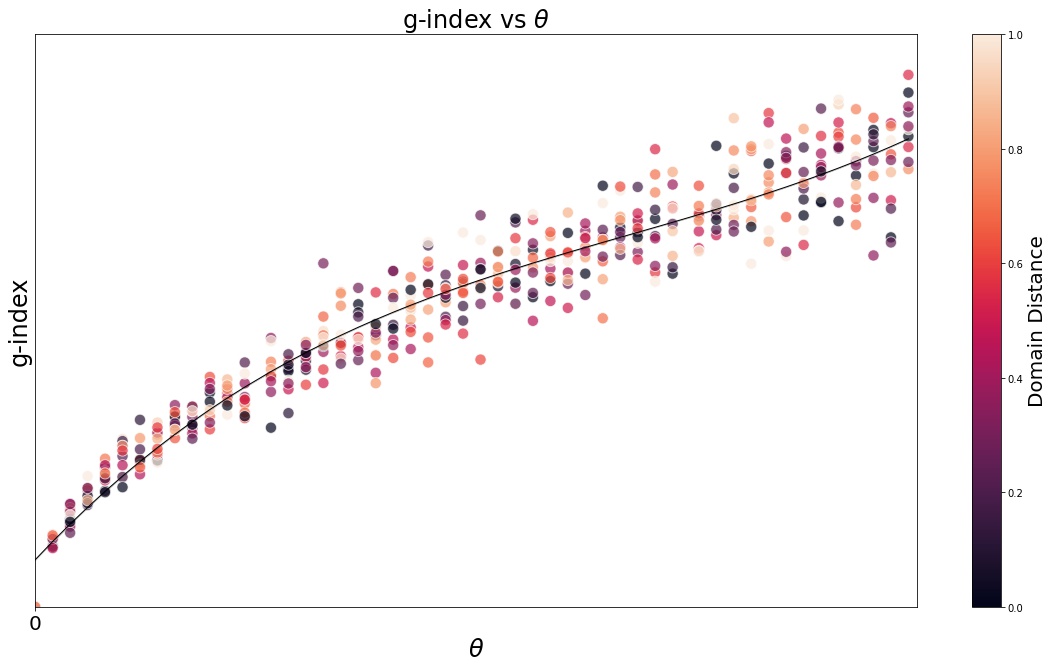}
	\caption{Responsiveness of the \gindex to an increase in performance of the system
		$\theta$. We assume the priors and experience of the system to fixed
		across all the points. The color of the points indicates the domain distance of tasks
		that the system was tested on -- darker points indicate the tasks were of low domain
		distance (the system did not display much generalization power), lighter points indicate the tasks were highly dissimilar from the curriculum
		$C$ of the system. We note that while the \gindex value increases
		as performance increases, generalization difficulty also plays a huge role, meaning systems that perform well on tasks with high generalization difficulty will have
		a high \gindex. We expect that the ideal intelligence system would be closer to the top
		right of the plot to achieve a high \gindex.}
	\label{fig:rando-graphs3}
\end{figure}

If the \textit{skill} of an intelligence system in a particular domain is
defined as it's ability to consistently generate a set of instructions (or programs) to
solve tasks in that domain, the ideal benchmark should aim to measure the efficiency of
acquiring new skills. It should penalise the amount of data and compute power required to
train the system, and should reward the performance of the system on tasks of increasing
generalization difficulty. If an intelligence system trained using the least amount of
training data and compute power obtains the highest performance on tasks of high
generalization difficulty, it should have the highest score on this benchmark. Keeping
these constraints in mind, we observe the responsiveness of the \gindex benchmark to
variations in the number of training samples (\autoref{fig:rando-graphs1}), compute power
(\autoref{fig:rando-graphs2}), performance and generalization difficulty
(\autoref{fig:rando-graphs3}) by running simulations across these variables.

\newpage
We see that the value of the \gindex decreases with an increase in the number of training
samples from \autoref{fig:rando-graphs1}, decreases with an increase in compute usage
from \autoref{fig:rando-graphs2}, and increases with an increase performance and
generalization difficulty from \autoref{fig:rando-graphs3}, which makes it useful for
measuring skill-acquisition efficiency.

\subsection{Levels of generalization}
\label{sub:levels-spectrum}

From our definition of the \gindex, we reason that \textit{general intelligence} is not a
binary property that a system either possesses or lacks, but is better described as a
continuous spectrum . Where an intelligence system lies on this spectrum depends on the
following factors of the evaluation setup:

\begin{itemize}
	\item The \textbf{diversity of the evaluation scope of tasks} $T'$ - whether or not they lie
	      within the similar domains where all tasks have low $\Delta$ divergence
	      score relative to each other.
	\item The \textbf{generalization difficulty} of tasks $T'$ with respect to
	      curriculum C, or how different the tasks in the test scope are from the curriculum it has
	      seen. We use the distance score $\Omega$ to refer to this.
	\item The \textbf{efficiency} with which a system converts its priors, experience with
	      curriculum $C$ to a high performance on the task scope
	      $T'$
\end{itemize}

We categorise intelligence systems into four levels of generalization power based on the
properties considered above. Each is harder to achieve than the previous one, and the
formulation of the \gindex makes it difficult to brute-force a higher score by utilising
unlimited amounts of priors, data and compute. The aim of these levels is to aid
subjective discussions about the strengths and weaknesses of different approaches to
build intelligence systems. We note that these levels merely represent approximate
demarcations of generalization difficulty; as the measurement of general intelligence
systems becomes more refined, these demarcations may change, or a new categorisation  may
be formulated.

\textbf{Level 0 - No generalization}. L0 broadly describes a system which encounters zero
uncertainty in the tasks on which it is evaluated. This is because all edge cases are
handled internally via hard-coded rules and the system does not act upon any learned
heuristic. For example, a sorting algorithm that outputs a sorted array every time in a
rule-based manner, or a chess playing algorithm that iterates through all possible moves
to win a game of chess, can be said to not display any generalization.

\textbf{Level 1 - Generalization to known variables in known domains}. An L1 general intelligence is able to generalize across
predictable amounts of uncertainty within a set of related tasks in the same domain.
Consider a set of task specifications of the form \textit{``Buy $X$ stock every $Y$ hours"}. The
variables $X$ and $Y$ here are
\textit{`known variables'}. In the program DAG $P'_{opt}$ for any of
these tasks, the variable $X$ maps to the
\texttt{tickername} attribute of the \texttt{submit-order} node type and
the variable $Y$ maps to the \texttt{frequency}
attribute of the \texttt{schedule-trigger} node type, both of which are wired together
(see \autoref{sec:node-types}). The degree of uncertainty is only in the attributes of
the nodes \texttt{submit-order} and \texttt{schedule-trigger}, hence the
generalization difficulty is low ( $\Bar{\Omega}\lesssim 0.15$). If an intelligence
system trained only on program DAG samples with different combinations of
$X$ and $Y$ is able to learn to generate
the correct program DAG for unknown combinations of $X$ and
$Y$, then it can be said to have reached L1 generalization.
Current deep learning-based approaches are successful at attaining this level.

\textbf{Level 2 - Generalization to unknown variables within known domains}. An L2 intelligence system is able to generalize to unknown
amounts of uncertainty within similar task domains. For example, when a system which has
been shown two different DAG programs for the task specifications
\textit{``Search for query on Google"}, and another to \textit{``Save a list of items to file"} is able to
successfully generate a program DAG for the new task specification
$T'$ : \textit{``Search for a query on Google and save results to file"}. This task has mid-range
generalization difficulty ( $0.4\lesssim\Bar{\Omega}\lesssim0.7$) because the system has to learn
how to combine two different program DAGs it has seen before. Unlike L1, the uncertainty
here is not just with the attributes of nodes, but also with the extra nodes and wires
needed to solve the task. This sort of composite synthesis within known domains can be
said an outcome of L2 generalization. While some deep learning systems occasionally
demonstrate this when exposed to large amounts of data and compute power, we expect that
sample-efficient methods of reaching L2 will require new approaches.

\textbf{Level 3 - Generalization to unknown variables across unknown domains}. An L3 intelligence system is able to adapt to arbitrary
amounts of uncertainty in the tasks on which it is evaluated. This is the most
challenging level because it requires the system to perform well on tasks with high
generalization difficulty ($\Bar{\Omega}\gtrsim0.85$), i.e. the program DAG required to
solve a task is highly dissimilar to any task it has seen before. For example, if an
intelligence system that is shown only web navigation tasks of the form
\textit{``Summarize page from Wikipedia.com"}, is asked to \textit{``Learn how to drive a Toyota on a given city street"}, it has to find
novel ways to convert its experience in one domain into mastery in another unknown
domain.        It could do this is by using its web navigation skills to watch an online
city-driving video tutorial, create a web-based simulation sandbox of the street with
virtual car controls, program new nodes to interface with the controls on a real car, and
then generate a program DAG to drive a car down a street. Current learning methods are
insufficiently equipped to create or scale up to such an L3 intelligence system. We
expect new methods will need to emerge which incorporate high sample-efficiency, embodied
agency, long-term memory and elements of self-improvement.

\section{Experiments}
\label{sec:expt}

In this section, we compute the values of the \gindex and its components for some
well-known large language models. We use a small dataset of text prompts and their
associated flow-based programs to finetune transformer-based models before measuring
their \gindex scores. We construct a small dataset of real-world tasks from
\textbf{16} task domains to train the models.  The task domains are
described in \autoref{sec:domains-full}. A sample of the dataset is available at \repo.
We finetune four transformer models: \textbf{GPT2-345M},
\textbf{GPT2-774M}, and \textbf{GPT2-1.5B} from
\citep{radford2019language}, and \textbf{GPT-Neo-2.7B} from
\citep{gpt-neo, gao2020pile}. We use the HuggingFace implementations
\citep{huggingface} of the transformer models in the experiments. \\

With the current set of experiments, we aim to measure skill-acquisition efficiency via
the \gindex with tasks of low generalization difficulty. The average domain distance
$\Omega$ between the training set and test sets across all experiments
is \textbf{0.09}.  The training samples range from
\textbf{640} to \textbf{10240} across all the experiments. In
every experiment, the training samples were distributed equally across all 16 task
domains. After training, the models are tested with \textbf{5}  unseen
samples per task to obtain their average performance $\theta$. In
every experiment, the number of training epochs was held constant at
\textbf{30}. When synthesizing the programs, we hold the
\texttt{temperature} of the models at a constant \textbf{0.7}, and
allow only one attempt at synthesis. We expect more attempts for a given task
specification will yield better performance \citep{copilot}. We examine the
following relationships: \\

\begin{figure}[htbp]
	\centering
	\includegraphics[width=0.95\linewidth,keepaspectratio]{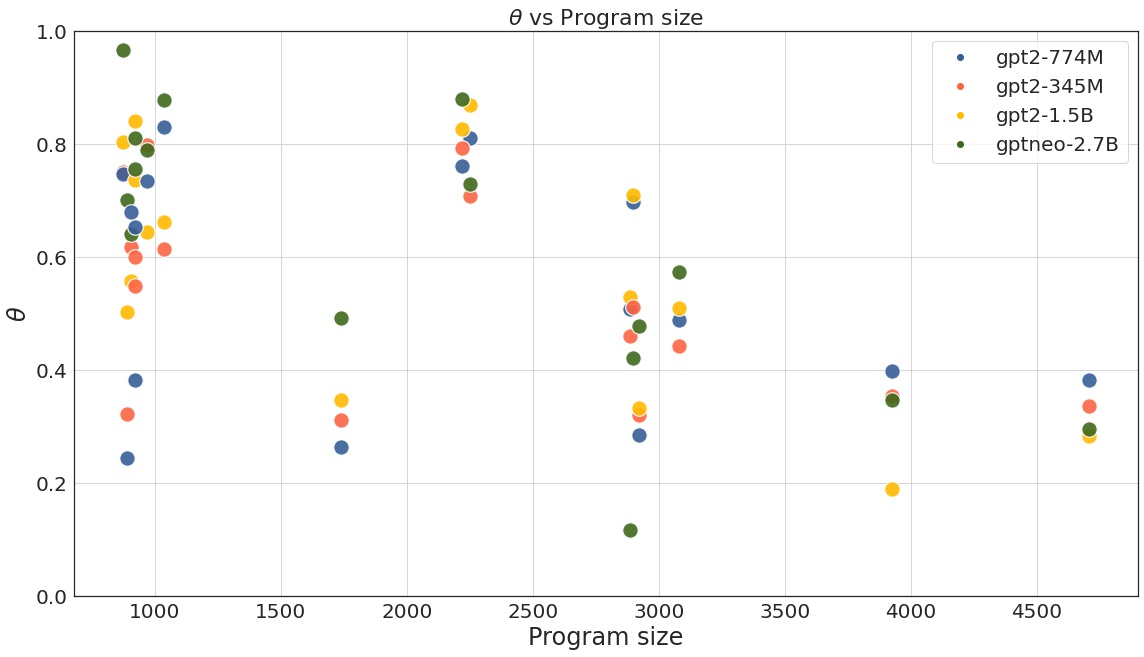}
	\caption{Average performance $\theta$ of each model as program size increases.
		The point color refers to a particular model (specified in the legend). Note that the
		decrease in $\theta$ is not uniform for all the
		models,
		suggesting that raw program length may not directly affect performance.
		However, the transformer models we use have a limited context window for
		generation, hence they falter in performance when the program size crosses a particular limit.}
	\label{fig:expt-graphs1}
\end{figure}

\begin{figure}[htbp]
	\includegraphics[width=0.95\linewidth,keepaspectratio]{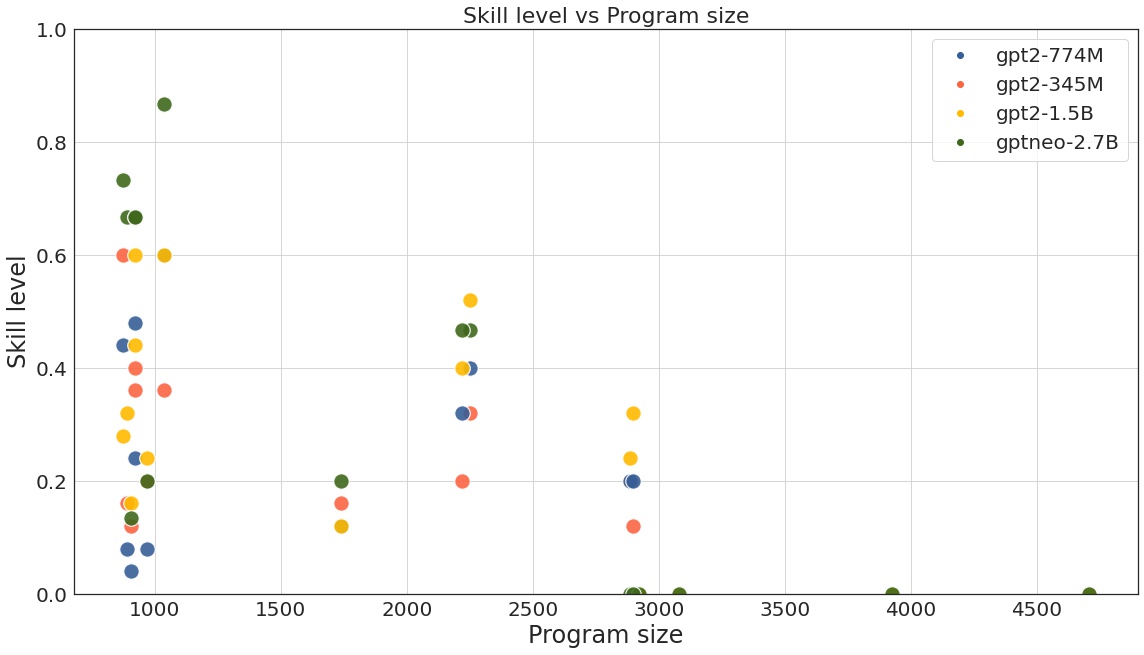}
	\caption{Skill level of each model as generated program size increases. The point color refers to
		a particular model (specified in the legend). The skill of the system is its ability to
		generate the correct program for a given task. For real-world use, we expect the systems to be able to generate the correct program in one attempt.
		However, we see that the skill levels of these models do not show that they can be used in
		real-world cases reliably just yet. The effect of program size is also more pronounced:
		beyond a certain limit, none of the models were able to produce a fully correct program for the specified task in
		one attempt. If the best out of multiple attempts is considered, we expect skill level to increase.
	}
	\label{fig:expt-graphs2}
\end{figure}

\begin{figure}[htbp]
	\centering
	\includegraphics[width=0.95\linewidth,keepaspectratio]{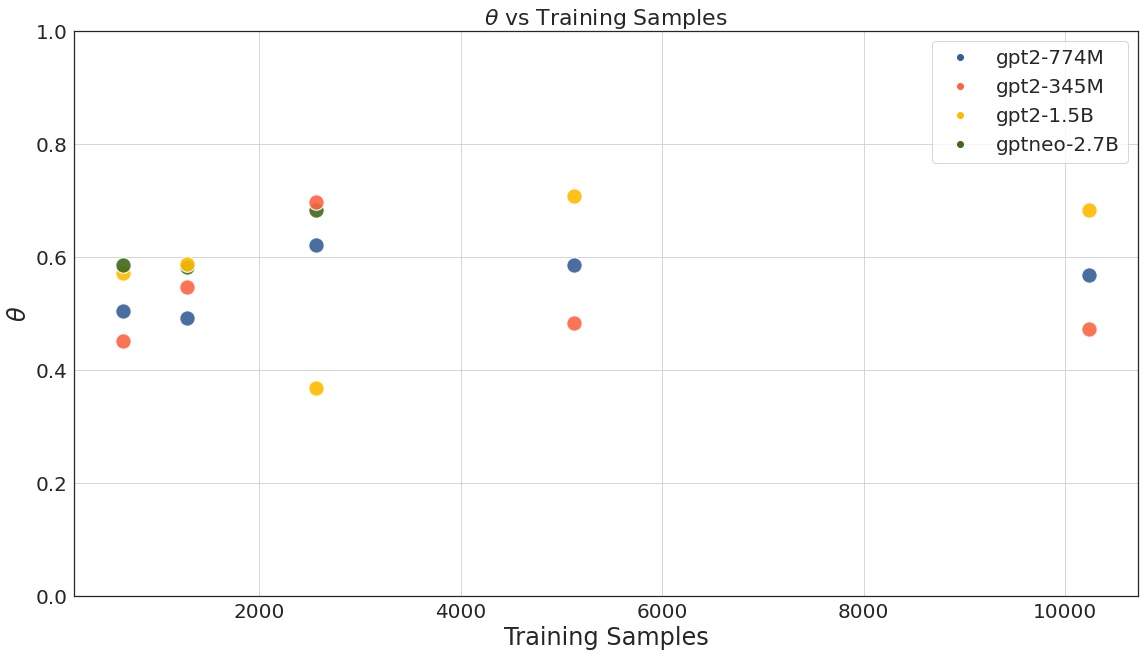}
	\caption{Average performance $\theta$ as the number of training samples
		increases. The bubble color refers to a particular model (specified in the legend). The
		bubble size is a relative measure of the \gindex score. The larger models require more
		samples to achieve better performance, but the increased performance does not offset the training samples penalty
		enough, and so their \gindex scores are lower.
	}
	\label{fig:expt-graphs3}
\end{figure}

\begin{figure}[htbp]
	\centering
	\includegraphics[width=0.95\linewidth,keepaspectratio]{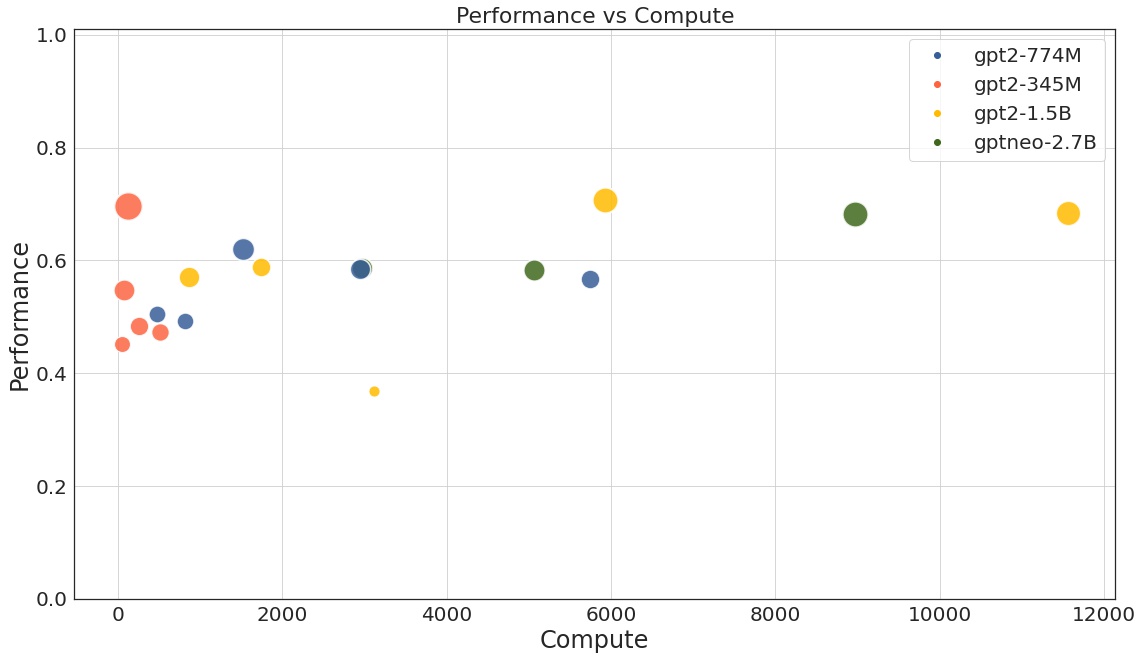}
	\caption{Average performance $\theta$ vs compute used. Compute used is measured
		as total peta floating point operations (total petaFLOPS used $\times$
		total training time in seconds). The bubble color refers to a particular model (specified
		in the legend), and the  bubble size is a relative measure of the \gindex score. The larger models take
		use far more compute, but the improvement in performance is marginal. This results in their \gindex scores
		being affected by to the compute penalty. Hence the models with the best performance may
		not have the best \gindex score.
	}
	\label{fig:expt-graphs4}
\end{figure}

\begin{enumerate}
	\item \textbf{average performance $\theta$ vs program size
		      (\autoref{fig:expt-graphs1})}: The programs generated for each task are of different
	      sizes. The size of the program (number of characters in the program text) affects how
	      easily it can be generated. For instance, the number of tokens transformer models can
	      generate is bounded by their context window. We expect model performance to falter as the
	      size of the program to generate increases.
	\item \textbf{skill levels vs program size (\autoref{fig:expt-graphs2})}: The skill of an intelligence
	      system is its ability to \textit{consistently} generate correct programs to solve
	      tasks in a given domain. In addition to performance, a measure of the system's skill in a
	      particular domain helps contextualize its potential for real-world use. When choosing an
	      intelligence system to deploy in real-world tasks, we would prefer a system that
	      generates correct programs more often.
	\item \textbf{average performance $\theta$ vs number of training samples
		      (\autoref{fig:expt-graphs3})}: The \gindex penalizes increments in training samples, but
	      rewards improvements in performance. We expect that the ideal system with a high \gindex
	      would occur at an optimal tradeoff between these two quantities.
	\item \textbf{average performance $\theta$ vs compute used
		      (\autoref{fig:expt-graphs4})}: The \gindex penalizes high compute usage, but rewards
	      improvements in performance. Compute is measured in terms of available compute power and
	      training time, so systems that use multiple processors in parallel are penalized
	      accurately. We expect the ideal system with a high \gindex would occur at an optimal
	      tradeoff between compute usage and performance.
\end{enumerate}

\begin{table}[htpb]
	\centering
	\renewcommand{\arraystretch}{1.5}
	\begin{tabular}{|c|c|c|c|c|c|}
		\hline
		   & Model Name   & \# Training Samples & Compute Used\footnote{f1} & $\theta$       & \gindex           \\ \hline
		1. & GPT2-345M    & 2560                & 127.530                   & 0.697          & 7902.972\supplus  \\ \hline
		2. & GPT Neo-2.7B & 2560                & 8969.100                  & 0.682          & 6421.049          \\ \hline
		3. & GPT2-1.5B    & 5120                & 5927.400                  & 0.708\supplus  & 6390.314          \\ \hline
		4. & GPT2-1.5B    & 10240\supplus       & 11563.320\supplus         & 0.683          & 6006.261          \\ \hline
		5. & GPT2-774M    & 2560                & 1516.640                  & 0.620          & 4872.334          \\ \hline
		6. & GPT Neo-2.7B & 1280\supminus       & 5063.380                  & 0.582          & 4476.680          \\ \hline
		7. & GPT2-345M    & 1280\supminus       & 74.750\supminus           & 0.547\supminus & 4399.190          \\ \hline
		8. & GPT2-774M    & 5120                & 2941.941                  & 0.585          & 4070.117\supminus \\ \hline

	\end{tabular}
	\vspace*{0.1in}
	\caption{The best 8 performing models sorted according to their \gindex scores. Values with
		\supplus indicate the maximum of the column, and values with \supminus indicate the minimum. We note that the system with the best
		$\theta$ is not the one with the highest
		\gindex, and the system with the worst $\theta$ score does not have the lowest
		\gindex. The differences between the top three rows indicate the tradeoffs involved:
		while row 3 has better performance with less compute, the increased performance does not offset the penalty of double the number of training
		samples.
	}

	\label{table:g-index-win}

	\vspace{-0.8em}
\end{table}

Different intelligence systems may obtain the best measurement at any individual
component of the \gindex. However, the best-performing system may not be the most
resource-efficient, and vice-versa. The ideal system would be one that has the right
combination of priors, experience, sample-efficiency, and maximal performance.
\autoref{table:g-index-win} shows the top 8 models according to their \gindex values,
along with the models' best component scores.

\footnotetext{Compute used is measured as total peta floating point operations (total petaFLOPS used
	$\times$ total training time in seconds) }

\section{Flatland - a toy environment for the \gindex}%
\label{sec:flatland}

While the evaluation setup in \autoref{sub:evalsetup} maps well to real-world tasks,
the development of new learning methods will require toy environments with simpler
program spaces and diverse task specifications. Since the \gindex rewards
sample-efficiency and performance on tasks of high generalization difficulty, new methods
that score a high \gindex within these environments will help surface insights for
real-world tasks. Recent efforts in program synthesis use visual examples as training
input to the system, such as images \citep{ellis2018learning, ellis2020dreamcoder} or video
\citep{sun2018neural, shu2021agent}. To examine the \gindex in such cases, we construct a toy
environment called \texttt{flatland}, where intelligence systems must infer the
correct program to draw shapes in an image.

\begin{figure}[htpb]
	\centering
	\hspace*{-0.2in}
	\includegraphics[width=1.1\linewidth]{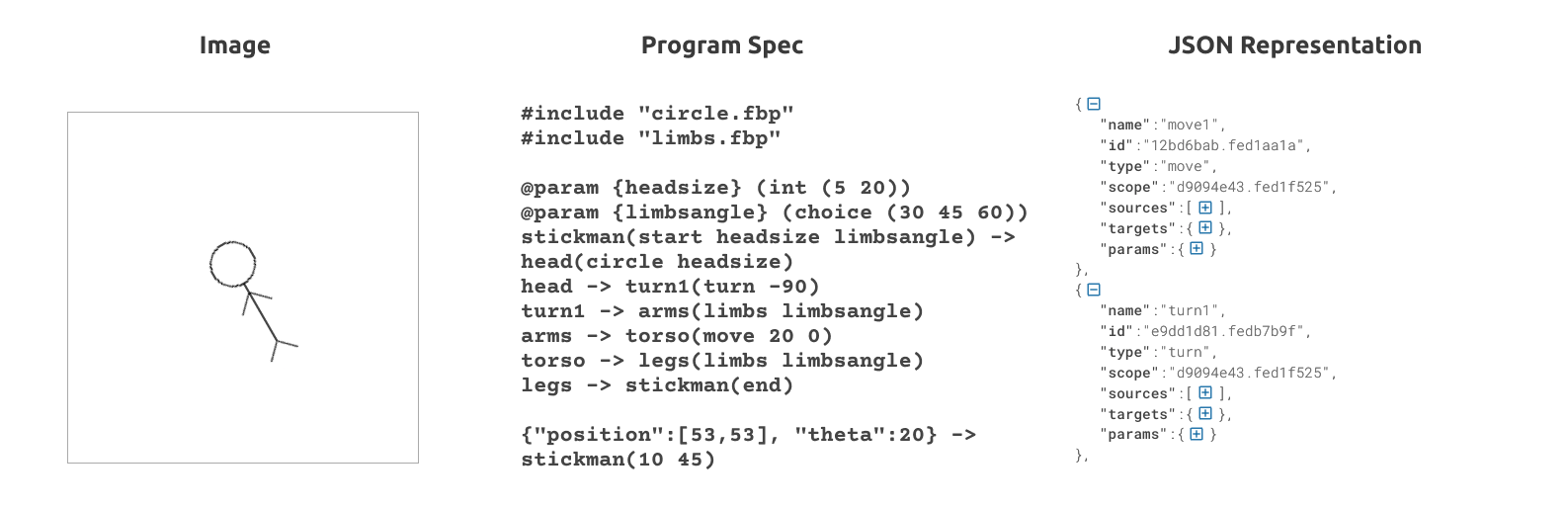}
	\caption{An example task in \texttt{flatland} with associated program and JSON
		representation. The JSON representation is similar to the flow-based programs for real-world tasks shown in
		\autoref{fig:ndr1}.}%
	\label{fig:flatland-sample}
\end{figure}

The evaluation setup for \texttt{flatland} is similar to \autoref{sec:factors} and is
described below\footnote{Code available at
\href{https://github.com/mayahq/flatland}{https://github.com/mayahq/flatland}} :

\begin{itemize}
	\item The \textit{task} $T'$ specified to the intelligence
	      system $IS$ is a 128x128 binary image. Each image contains shapes
	      drawn using the commands \texttt{loop}, \texttt{move}, and \texttt{turn} as primitives.
	\item The \textit{program spec} $P'$ generated by $IS$ can be expressed in a
		flow-based syntax, LISP expressions, or as JSON similar to programs for real-world tasks.  \autoref{fig:flatland-sample}
		shows an example. We use  the \texttt{turtle} graphics library in Python
		\citep{pyturtle} (similar to LOGO \citep{logo1972}) to draw the images from the
		DAG program specification. The DAG program is converted into a JSON list of
		primitives for comparison.
	\item The curriculum $C$ for $IS$ consists of images $T$ and the corresponding DAG programs
	      $P$ that generate them. \texttt{flatland} has in-built data augmentation to create programs with
		slight variations from an existing program. 
	  \item The \textit{scoring function} evaluates the output of $IS$ to
	      the ideal program $P'$ with a match-based metric
	      $\Delta$ that calculates the largest common subsets between the two
	      lists. The largest common subset is computed via an association graph, similar to
	      \autoref{sub:scoring}.
	\item The \textit{generalization difficulty} $GD(T', C)$ of a task
	      $T'$ with the ideal program $P'$ is
	      computed by finding the nearest neighbor of $P'$ in
	      $C$, similar to \autoref{sub:measure-gd}.
	      \autoref{fig:flatland-omega} shows some test samples along with their domain distance
	      $\Omega$ from a curriculum $C$ of just lines
	      and circles.
	      \begin{figure}[htpb]
		      \centering
		      \includegraphics[width=1\linewidth]{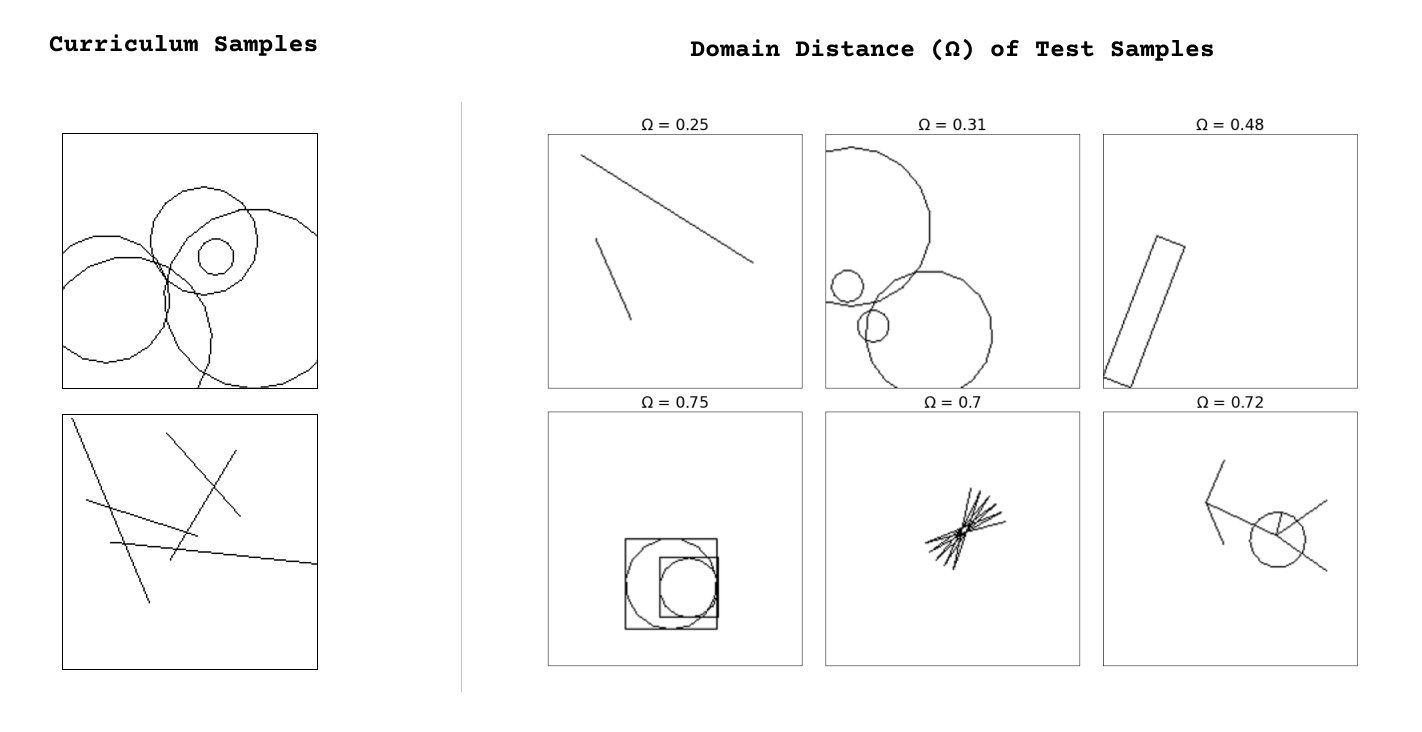}
		      \caption{Samples from the training and test sets for \texttt{flatland}. The training set
			      consists of various images containing one to five circles or lines. Above each test
			      sample we display its domain distance $\Omega$ from the training set.}%
		      \label{fig:flatland-omega}
	      \end{figure}
\end{itemize}

With \texttt{flatland}, we wish to examine intelligence systems by their \gindex values in
a controlled environment, so that we can test which learning methods generalize well when
some constraints are relaxed. We also hope to improve the \gindex definition by varying
the evaluation setup itself.  In the \texttt{flatland} environment, every aspect of
program synthesis can be controlled and tested:

\begin{itemize} 
	
	\item The \texttt{flatland} environment currently provides programs that produce
	images, but this can be changed to accompany the image with text, or even use video.
		The program can be expressed in a flow-based syntax, or with LISP expressions, or
		also as JSON. This can help test synthesis methods that depend on different
		modalities.

	\item The DAG programs in \texttt{flatland} currently consist of three primitives
		\texttt{loop}, \texttt{move}, and \texttt{turn}. By varying the base set of
		primitives, we can attempt to find an ideal combination with which a system can
		learn to express the widest range of programs.

	\item The primitives in \texttt{flatland} can be combined to draw shapes, which can
		further be combined to produce more complex drawings. Thus new programs can be
		created via reuse and composition of existing programs, while keeping the program
		text minimal. This hierarchical composability could help create more flexible
		learners which build concepts on top of existing concepts to synthesize programs.

	\item \texttt{flatland} has in-built data augmentation to create programs with slight
		variations from an existing program, allowing the generation of arbitrarily large
		datasets on demand. As \texttt{flatland} programs are structurally similar to
		those used for real-world tasks (see \autoref{fig:flatland-sample} and
		\autoref{fig:ndr1}), this could help assess the format, quantity, and diversity of
		training data that will be needed if a system has to adapt to tasks in the real
		world.  

\end{itemize}

\section{Summary and Future Directions}
\label{sec:conclusion}

In this paper, we describe an experiment framework to obtain a quantitative measurement
of skill-acquisition efficiency of machine intelligence systems. We model the
intelligence system to accept a wide range of real-world tasks specified in natural
language and synthesize programs representable as directed acyclic graphs in a flow-based
programming syntax. We define a match-based metric to compare the DAG structures of two
given programs to score the performance of the system, and use this metric to measure the
generalization difficulty of any task provided to the system. We formulate the \gindex
benchmark and show that its changes with respect to dataset size, available compute, and
performance align with intuitive expectations of an intelligence system with high
generalization power. We then measure and compare the \gindex scores of some fine-tuned
transformer models and estimate their suitability for general-purpose intelligence
systems. Finally, we provide a toy environment with a relaxed set of constraints to
examine potential designs and improvements to the \gindex.

While the \gindex benchmark shifts the evaluation of intelligence systems into a
quantitative context akin to skill-based evaluations, it is not yet a complete measure of
general machine intelligence. However, we believe that future evaluations of intelligence
systems will require a similar framework, one that reflects the potential real-world use
of such systems. Over the course of our experiments, we have found some possible
directions for improving the \gindex measurement. We describe these possibilities and
their effects below.

\textbf{The evaluation framework} can be improved to represent more real-world use cases
for a machine intelligence system.	The task specification can be expanded to include
different natural languages, audio, video, and input-output examples. The task may even
be specified in parts across an interaction between a human and the system
\citep{austin2021program}.  Flow-based programs face limitations with maintaining
state, so there is potential for improving the language of synthesized programs to
account for multi-stage tasks. It is also possible to create more nodes with new
functionality, enabling the construction of larger, diverse datasets of task
specifications and their associated programs. With larger datasets, better data
augmentation techniques can be built to generate reference programs to evaluate tasks. As
more nodes are designed and flow-based programs grow larger, the subgraph isomorphism
computation with $\Delta$ may slow down, and the chance of program
aliasing issues may also increase. In such cases, we may also need to use functional
correctness like \citep{copilot} to score the programs synthesized by the
system.

With a given a set of tasks, the \textbf{components of the g-index} such as compute and domain
distance can be refined by testing with a wider variety of intelligence systems, to
ensure that the \gindex value accurately represents their capabilities. Additionally,
since the intelligence systems in our framework are evaluated on real-world tasks,  human
feedback can be incorporated when attempting to understand or improve the calculation of
these components.

After calculating the scores for a wide range of systems, \textbf{the g-index formula}
would need to be updated with additional specifications. If a system's \gindex score is
unnaturally high, perhaps the system actually exhibits high skill-acquisition efficiency,
or the formula contains some poorly specified variables which unfairly portray the
system's ability. For example, we consider the weightage for the priors encoded in the
system ($\rho$ in \autoref{eq:task-contr}) to be a constant
neglible value for the current set of experiments that use transformer models fine-tuned
from pre-trained weights.  When comparing the \gindex of a fine-tuned model to a model
trained from scratch, the priors/compute tradeoff would be different, but it is not clear
how such differences can be measured. Going further, it is difficult to quantify the
benefit of priors like hyperparameters, data preprocessing, hardcoded rules and initial
primitives in a manner that translates fairly across different kinds of intelligence
systems. Perhaps the weight of some priors $P_k$ can be calculated
by comparing the system's performance with and without the prior:

\begin{equation}
	\label{eq:prior-hax}
	\rho(P_k) \propto \frac{\text{\gindex}(IS \text{~with prior~} P_k, T')}{\text{\gindex}(IS \text{~without prior~} P_k, T')}
\end{equation}

The arrangement of \autoref{eq:prior-hax} along with the levels of generalization
discusssed in \autoref{sub:levels-spectrum} leads to an interesting question. When
comparing machine intelligence with human intelligence, given that human beings have had
thousands of years to build priors, should human priors have a weight of
$\rho_{\text{human}} \rightarrow \infty$? Is it necessary to compare the built-in priors of humans
with that of machines?

The overarching aim of this work is not only to propose a method-agnostic way to compare
different techniques quantitatively, but also to spark a conversation on the relative
merits of different approaches that could help reach higher levels of general machine
intelligence. While we don't expect our \gindex definition to be complete, it anchors a
previously abstract concept in a mathematical formulation made up of quantities that can
be measured during experiments. The flow-based programming language we propose can act as
a common language to express and compare programs of real-world utility across a wide
variety of domains. The \gindex explicitly rewards resource-efficiency, which we hope
incentivises sustainable ways of achieving generalization which do not rely on unlimited
amounts of compute and data. A good outcome of this would be the patronage and
competitive development of new methods and technologies to reach higher \gindex levels,
similar in spirit to the Hutter compression challenge \citep{hutter-prize}.
Ultimately, our belief is that intelligent machines will only be able to contribute to
real technological progress when they can learn how do \textit{more from less}, not
the other way around.

\newpage

\printbibliography

\appendix

\savegeometry{default}
\newgeometry{,hmargin=1.0cm,vmargin=0.5cm}
\pagestyle{empty}
\section{Sample Task Domains and Descriptions}
\label{sec:domains-full}
\renewcommand{\arraystretch}{2.0}
\begin{table}[htbp]
	\centering
	\begin{tabular}
		{|p{25mm}|p{75mm}|p{10mm}|p{15mm}|p{20mm}|p{15mm}|}
		\hline
		Domain Name         & Example Text Prompt                                                                                                                                                                                             & Nodes & Program size(chars) & Best Avg Performance & Skill Level ** \\ \hline
		template-slider     & Two sliders with values ranging from 0 to 20 changing in steps of 2                                                                                                                                             & 6     & 1493                & 1.00                 & 1.00           \\ \hline
		template-form       & Create a form with fields for entering Name, Benchmark Score and Date of Submission                                                                                                                             & 5     & 1473                & 1.00                 & 1.00           \\ \hline
		cron-reminder       & Set a reminder for 'Send Daily Digest' every first day of the week, Tuesday through Saturday, only in January                                                                                                   & 3     & 935                 & 1.00                 & 1.00           \\ \hline
		cron-schedule       & Repeat At 48, 9, 14, 7, 6, 56, 46, 39, 15, 3, 37, and 30 minutes past the hour, between 05:00 AM and 08:59 PM, on day 1,4,8 and 9 of the month, only on Tuesday, every 3 months, November through December      & 3     & 896                 & 1.00                 & 1.00           \\ \hline
		facebook            & On Facebook, when user says 'Hello', reply with 'Hi there! How can I help you?', and when user says 'Bye', reply with 'Goodbye!'                                                                                & 8     & 2253                & 1.00                 & 1.00           \\ \hline
		gmail-send          & Send email with body 'Upcoming Meeting' and subject 'Discuss paper appendices' to email alpha@beta.com                                                                                                          & 12    & 4708                & 0.63                 & 0.00           \\ \hline
		google-search       & Search Google for 'How to make tables on LaTeX' and scrape results                                                                                                                                              & 9     & 2963                & 0.60                 & 0.0            \\ \hline
		googlesearch-to-csv & Search Google for "Papers on measuring general intelligence", scrape results and put into singularity.csv                                                                                                       & 12    & 3894                & 0.59                 & 0.0            \\ \hline
		http                & Create a HTTP POST endpoint called /agents                                                                                                                                                                      & 4     & 857                 & 1.0                  & 1.0            \\ \hline
		telegram-2-reply    & Reply 'Yes, detective?' to 'Sonny!', and 'Of course, Dave', to 'Open the pod bay doors' on Telegram                                                                                                             & 8     & 2148                & 1.0                  & 1.0            \\ \hline
		telegram-3-reply    & On Telegram, when user says 'Are friends electric?', reply with 'No, only sheep', when user says 'How deep is your love?' reply with '6.5 meters', and when user says 'Is that all there is?', reply with 'Yes' & 10    & 2910                & 1.0                  & 1.0            \\ \hline
		twitter             & Obtain tweets about the topic \#Alignment                                                                                                                                                                       & 4     & 892                 & 1.0                  & 1.0            \\ \hline
		url-skill           & Create a skill called 'Open LessWrong' which opens url https://www.lesswrong.com/                                                                                                                               & 4     & 1067                & 1.0                  & 1.0            \\ \hline
		youtube-pause       & Pause Youtube video                                                                                                                                                                                             & 3     & 919                 & 1.0                  & 1.0            \\ \hline
		youtube-play        & Find and play Vivaldi Four Seasons on Youtube                                                                                                                                                                   & 9     & 3050                & 0.89                 & 0.0            \\ \hline
		youtube-resume      & Resume Youtube Video                                                                                                                                                                                            & 3     & 916                 & 1.0                  & 1.0            \\ \hline
	\end{tabular}
	\vspace*{0.1in}
	\caption{Task domains consist of DAGs with mean $\Omega <$
		0.1, i.e. with different types of known variables}
	\label{tab:domain-descr}
\end{table}

\newpage
\section{Node Library}
\label{sec:node-types}
\renewcommand{\arraystretch}{1.8}
\begin{table}[htbp]
	\centering
	\begin{tabular}
		{|p{35mm}|p{65mm}|p{70mm}|}
		\hline
		Categories                  & Description                                                                                    & Node Types                                                                                                                                    \\ \hline
		\textbf{common utility}     & Custom triggers, catch bugs, add comments                                                      & \texttt{inject, debug, complete, catch, status, link in, link out, comment}                                                                   \\ \hline
		\textbf{functional}         & Change, switch, filter or delay the passed message object or add custom logic to manipulate it & \texttt{function, switch, change, range, template, delay, trigger, exec, filter}                                                              \\ \hline
		\textbf{network}            & Different kinds of network interfaces to send and receive data                                 & \texttt{mqtt in, mqtt out, http in, http response, http request, websocket in, websocket out, tcp in, tcp out, tcp request, udp in, udp out}  \\ \hline
		\textbf{sequence}           & Manipulate sequences and arrays in predictable ways                                            & \texttt{split, join, sort, batch}                                                                                                             \\\hline
		\textbf{parser}             & Parse data from different files into fixed formats for easy processing                         & \texttt{csv, html, json, xml, yaml}                                                                                                           \\ \hline
		\textbf{storage}            & Read and write to files                                                                        & \texttt{file, file in, watch}                                                                                                                 \\ \hline
		\textbf{dashboard}          & Make dynamic dashboards with forms and charts by linking UI elements to other pieces of logic  & \texttt{button, dropdown, switch, slider, numeric, text input, date picker, colour picker, form, text, gauge, chart, audio out, notification} \\ \hline
		\textbf{browser-automation} & Interact with the browser to navigate the web and scrape websites                              & \texttt{open, click, type, press, execute function, find tab, scrape, query, bookmark}                                                        \\ \hline
		\textbf{spotify-automation} & Integrate with spotify and control music played on any device                                  & \texttt{play, search, control playback, get playback state, control playlist}                                                                 \\ \hline
		\textbf{gdrive-automation}  & Search through, read, export and append to files in on your google drive                       & \texttt{search gdrive, gsheet append, gdrive-export-file}                                                                                     \\ \hline
		\textbf{scheduling}         & Schedule triggers to run events at any fixed interval                                          & \texttt{schedule-trigger}                                                                                                                     \\ \hline
		\textbf{zoom-automation}    & Create, view and attend zoom meetings                                                          & \texttt{create-meeting, list-meetings, list-meetings-registrants}                                                                             \\ \hline
		\textbf{system utilities}   & Interact with various system level utilities on the desktop                                    & \texttt{clipboard-add, clipboard-get, open-target, file-search, desktop-notify}                                                               \\ \hline
		\textbf{stock-automation}   & Buy, sell and view orders on the stock market via third party API                              & \texttt{submit-order, get-order, get-bars, get-account}                                                                                       \\ \hline
		\textbf{home-automation}    & Control lights and switches remotely                                                           & \texttt{light-control, switch-control}                                                                                                        \\ \hline
	\end{tabular}
	\vspace*{0.1in}
	\caption{Each node is a reusable, encapsulated "black box" function that can be wired to other nodes to form a program DAG
		and automate any task.}
	\label{tab:nodetypes-descr}
\end{table}

\loadgeometry{default}
\section{Calculating $\Delta$ via subgraph isomorphism}
\label{sec:delta-actual}

When computing the value of $\Delta$ between two arbitrary DAGs, we
use the similarity function $w$ to compare individual nodes, and
account for structural similarity by computing the largest subgraph common to both DAGs.
This is known as the maximum common edge subgraph problem, \citep{bokhari1981mapping},
an extension of the subgraph isomorphism problem \citep{ullmann1976algorithm}. In this
section, we explain in detail the calculation of $\Delta$ outlined in
\autoref{ssub:div-everything}.\\

Two graphs $G'(V', E')$ and $G''(V'', E'')$ are isomorphic to
each oher ($G' \simeq G''$) if there exists a bijection
$\phi: V'  \rightarrow V''$ that preserves the graph structure:

\begin{equation}
	(v_1, v_2) \in E' \iff (\phi(v_1), \phi(v_2)) \in E'' ~~~ \forall (v_1, v_2) \in E'
	\tag{\ref{eq:iso-struct}}
\end{equation}

We need to find the largest subgraphs $ G'_{opt} \subseteq G' $ and
$ G''_{opt} \subseteq G'' $ that are isomorphic to each other
($G'_{opt} \simeq G''_{opt}$), and the isomorphism $\phi_{opt}$ between
the nodes of the two subgraphs. We obtain these by constructing the
\textit{association graph} $G$ and finding a maximum clique in
$G$. The association graph $G(V,E)$ for given
graphs $G'$, $G''$ contains the vertices
$V \subset V' \times V'' \times (0, 1]$: a vertex $(v', v'', w) \in V$ associates a node
$v' \in V'$ to a node $v'' \in V''$ with their node
similarity $w(v', v'')$. Only nodes of the same
\texttt{type} are considered for association:

\begin{equation}
	\label{assoc-verts}
	(v'_i, v''_i, w_i) \in V \iff w_i = w(v'_i, v''_i) > 0
\end{equation}

An edge in $G$ connects a vertex $(v'_1, v''_1, w_1)$ to
another vertex $(v'_2,
	v''_2, w_2)$ provided the below \textit{structure-preserving}
condition is satisfied (note the resemblance to \autoref{eq:iso-struct}):

\begin{equation}
	\label{assoc-edges}
	\begin{aligned}
		((v'_1, v''_1, w_1), (v'_2, v''_2, w_2)) \in E & \iff &  & ((v'_1, v'_2) \in E' \iff (v''_1, v''_2) \in E'')
	\end{aligned}
\end{equation}

It can be shown that finding a maximum clique\footnotemark in the association graph
$G$ is equivalent to finding the largest common subgraph between
$G'$ and $G''$ \citep{barrow1976subgraph, kozen1978clique}.
Once a maximum clique in $G$ has been obtained, we can construct
the common subgraphs $G'_{opt}$ and $G''_{opt}$. Since
nodes in the DAGs $G'$ and $G''$ cannot have
self-loops, \autoref{assoc-edges} ensures that any clique in the association graph
$E$ will always provide a one-to-one mapping between the
corresponding vertices of $G'$ and $G''$. If
$$ C_{opt} = \{(v'_1, v''_1, w_1), (v'_2,
	v''_2, w_2) ... (v'_N, v''_N, w_N)\}$$

\footnotetext{
	Note that the node similarities $w_i$ are used to filter out
	elements from $V$ and when computing the maximum clique (a
	node-weight maximum clique is computed).
}

is the maximum clique in the association graph $G$, then

$$ G'_{opt} = G'(V'_{opt},~\text{the subgraph of $G'$ induced from} ~~ V'_{opt} = \{v'_1, v'_2 ... v'_N\}
	\text{~and~}$$
$$ G''_{opt} = G''(V''_{opt}),~\text{the subgraph of $G''$ induced from} ~~ V''_{opt} =\{v''_1, v''_2 ...
	v''_N\} $$

are the required largest common subgraphs. The maximum subgraph isomorphism
$\phi_{opt}$ is the mapping expressed via the pairs
$(v'_i, v''_i)$ in the elements of of the clique $C_{opt}$.

$$ \phi_{opt} : V'_{opt} \rightarrow V''_{opt} \implies
	\phi_{opt}(v'_i)
	= v''_i,
	\text{~where~} (v'_i, v''_i, w(v'_i, v''_i)) \in C ~~
	\forall~v'_i \in V'_{opt}
$$

Thus with $C_{opt}, G'_{opt}, G''_{opt}, \text{~and~}
	\phi_{opt}$ we can compute the value of the metric
$\Delta(G', G'')$ as provided in \autoref{metric-full-glitch}:

\begin{align}
	\label{metric-full}
	\Delta(G', G'') & = 1 - \frac{(\sum^{N}_{i=1}w(v'_i, \phi_{opt}(v'_i))^2}{|V'||V''|}  \tag{\ref{metric-full-glitch}} \\\nonumber
	\Delta(G', G'') & = 1 - \frac{(\sum^{N}_{i=1}w(v'_i, v''_i)^2}{|V'||V''|}, (v'_i, v''_i, w(v'_i, v''_i) = w_i) \in C \\
	                & = 1 - \frac{(\sum^{N}_{i=1}w_i ^2}{|V'||V''|}
\end{align}

We note that $\Delta$ is 1 when either of the DAGs are empty, and
reduces to the node similarity function $w$ in
\autoref{metric-simple} when $G', G''$ both have one vertex and
zero edges. $\Delta$ also satisfies the constraints in
\autoref{metric-constraints}.

\begin{itemize}
	\item $\Delta$ is bounded:
	      \begin{equation}
		      \begin{split}
			               & w_i \in [0,1]~\forall~w_i                                                                          \\
			               & \sum^{n}_{i=1}w_i = |V'_{opt}| = |V''_{opt}|  \iff w_i = 1~\forall~w_i                             \\
			      \implies & \sum^{n}_{i=1}w_i \leq |V'_{opt}| = |V''_{opt}|                                                    \\
			               & |V'| = |V'_{opt}| = |V''_{opt}| = |V''| \iff G' \simeq G''                                         \\
			      \implies & |V''_{opt}| \leq |V'|, |V''_{opt}| \leq |V''|                                                      \\
			      \implies & 0 \leq \frac{(\sum^{n}_{i=1}w_i)^2}{|V'||V''|} \leq \frac{|V'_{opt}||V''_{opt}|}{|V'||V''|} \leq 1 \\
			      \implies & \Delta(G', G'') \in [0,1]~~\forall~~G', G''
		      \end{split}
	      \end{equation}

	\item $\Delta$ is symmetric because the node similarities
	      $w_i$
	      are symmetric, and finding the largest common subgraph between two graphs is also
	      symmetric: since $G'_{opt} \simeq G''_{opt}$, we can construct a bijection
	      $\phi$ between $G'_{opt}$ and
	      $G''_{opt}$, and use its inverse $\phi^{-1}$ for the
	      symmetric case.

	\item If $\Delta(G', G'') = 1$,

	      \begin{equation}
		      \begin{split}
			      \Delta(G', G'') = 1 & \implies \frac{(\sum^{n}_{i=1}w_i)^2}{|V'||V''|} = 0                  \\
			                          & \implies  w(v', v'') = 0~~\forall v' \in V', \forall v'' \in V'       \\
			                          & \implies |G'_{opt}| = |G''_{opt}| = |C_{opt}| = 0                     \\
			                          & \implies  \text{no clique $C$ could be found in the association graph
			      $G(V,E)$}                                                                                   \\
			                          & \implies  |G| = 0 \text{,~~ $G$ contains no vertices}                 \\
			                          & \implies  G', G'' \text{~~~do not have any nodes in common}
		      \end{split}
	      \end{equation}

	\item If $\Delta(G', G'') = 0 $,
	      \begin{equation}
		      \begin{split}
			      \Delta(G', G'') = 0                                           & \implies    (\sum^{n}_{i=1}w_i)^2 = |V'||V''|                \\
			      \text{but}~~~ (\sum^{n}_{i=1}w_i) \leq |V'_{opt}| \leq |V'|   & \implies    (\sum^{n}_{i=1}w_i) = |V'_{opt}| = |V'|          \\
			      \text{but}~~~ (\sum^{n}_{i=1}w_i) \leq |V''_{opt}| \leq |V''| & \implies    (\sum^{n}_{i=1}w_i) = |V''_{opt}| = |V''|        \\
			                                                                    & \implies  w_i = 1~~\forall (v'_i, v''_i, w_i) \in C_{opt}    \\
			                                                                    & \implies |V'| = |V'_{opt}| = |C_{opt}| = |V''_{opt}| = |V''| \\
			                                                                    & \implies  G' =  G'_{opt} \simeq G''_{opt} = G''              \\
			                                                                    & \implies \text{the two DAGs denote the same program}
		      \end{split}
	      \end{equation}
\end{itemize}

\end{document}